%% file: main.tex
\documentclass[letterpaper]{article} 
\usepackage{aaai25}  
\usepackage{times}  
\usepackage{helvet}  
\usepackage{courier}  
\usepackage[hyphens]{url}  
\usepackage{graphicx} 
\usepackage{natbib}  
\usepackage{caption} 
\usepackage{algorithm}
\usepackage{algorithmic}

\usepackage{amsmath}
\usepackage{multirow}
\usepackage{xspace}
\usepackage{threeparttable}
\usepackage{url}
\usepackage{diagbox}
\usepackage{colortbl}
\usepackage{makecell}
\usepackage{caption}
\usepackage{booktabs}
\usepackage{subfig}
\usepackage{graphicx}
\usepackage{xcolor}         
\usepackage{amssymb}
\usepackage{enumitem}
\usepackage{newfloat}
\usepackage{listings}
\DeclareCaptionStyle{ruled}{labelfont=normalfont,labelsep=colon,strut=off} 
\floatstyle{ruled}
\newfloat{listing}{tb}{lst}{}
\floatname{listing}{Listing}
\newcommand{\ie}{\textit{i}.\textit{e}.}
\newcommand{\eg}{\textit{e}.\textit{g}.} 
\newcommand{\wrt}{\textit{w}.\textit{r}.\textit{t}}
\def\model{UrbanVLP\xspace}
\def\dataset{CityView\xspace}

\title{UrbanVLP: Multi-Granularity Vision-Language Pretraining for \\  Urban Socioeconomic Indicator Prediction}
\author {
    Xixuan Hao\textsuperscript{\rm 1}\equalcontrib,
    Wei Chen\textsuperscript{\rm 1}\equalcontrib,
    Yibo Yan\textsuperscript{\rm 1}, 
    Siru Zhong\textsuperscript{\rm 1}, 
    Kun Wang\textsuperscript{\rm 2}, 
    Qingsong Wen\textsuperscript{\rm 3}, 
    Yuxuan Liang\textsuperscript{\rm 1}\thanks{Corresponding author.}
}
\affiliations {
    \textsuperscript{\rm 1}The Hong Kong University of Science and Technology (Guangzhou), Guangzhou, China\\
    \textsuperscript{\rm 2}National University of Singapore, Singapore\\
    \textsuperscript{\rm 3}Squirrel AI, Bellevue, USA\\
    
    \{xhao390, wchen110, szhong691\}@connect.hkust-gz.edu.cn; \\
    \{yanyibo70, qingsonedu\}@gmail.com; wk520529@mail.ustc.edu.cn; yuxliang@outlook.com
}

\usepackage{bibentry}

\begin{document}

\maketitle

\begin{abstract}
Urban socioeconomic indicator prediction aims to infer various metrics related to sustainable development in diverse urban landscapes using data-driven methods. However, prevalent pretrained models, particularly those reliant on satellite imagery, face dual challenges. Firstly, concentrating solely on macro-level patterns from satellite data may introduce bias, lacking nuanced details at micro levels, such as architectural details at a place.   
Secondly, the text generated by the precursor work UrbanCLIP, which fully utilizes the extensive knowledge of LLMs, frequently exhibits issues such as hallucination and homogenization, resulting in a lack of reliable quality.
In response to these issues, we devise a novel framework entitled UrbanVLP based on Vision-Language Pretraining. Our UrbanVLP seamlessly integrates multi-granularity information from both macro (satellite) and micro (street-view) levels, overcoming the limitations of prior pretrained models. 
Moreover, it introduces automatic text generation and calibration, providing a robust guarantee for producing high-quality text descriptions of urban imagery.
Rigorous experiments conducted across six socioeconomic indicator prediction tasks underscore its superior performance.  

\end{abstract}

\begin{links}
    \link{Code}{https://github.com/CityMind-Lab/UrbanVLP}

\end{links}

\vspace{-1em}
\section{Introduction}
\emph{Urban Socioeconomic Indicator (USI) Prediction}, a scholarly pursuit in the realm of urban computing~\cite{zou2024deep}, deploys data-driven methodologies to forecast socioeconomic metrics (e.g., GDP, population, carbon emission).
Rooted in the escalating global prominence of urban environments and the imperative for judicious urban planning, this academic discipline finds motivation in enhancing the efficacy of policy-making, optimizing resource allocation, and mitigating challenges endemic to sustainable development~\cite{xi2022beyond,yan2023urban}.

\begin{figure}[!t]
    \centering
    \includegraphics[width=\linewidth]
    {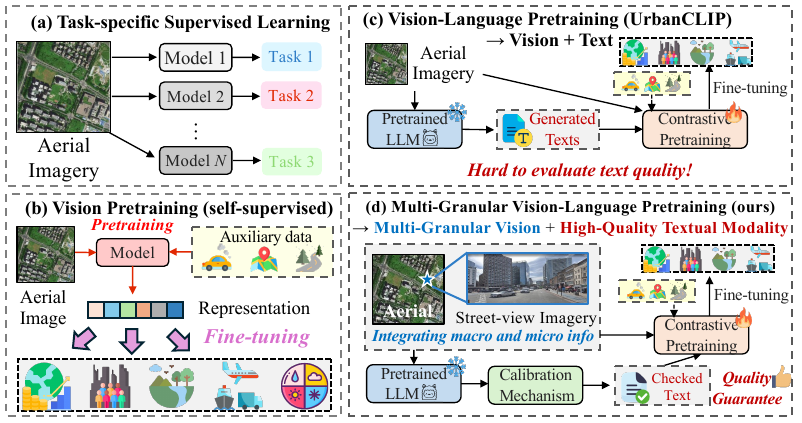} 
    \caption{USI prediction frameworks. Compared to existing arts, we present the first attempt to introduce multi-granular visual information and high-quality calibrated texts.}

    \label{fig:intro}
    \vspace{-1.2em}
\end{figure}

Given the widespread accessibility of satellite imagery on platforms such as Google Maps~\cite{li2022predicting,liu2023knowledge,xi2022beyond}, coupled with its wealth of information on regional features (\emph{e.g.}, road networks, building density, and vegetation coverage), a predominant majority of initiatives \emph{harness satellite imagery as the foundational modality} for learning urban region representations and making predictions~\cite{zhang2024towards}.
Through a comprehensive review of the current literature, we summarize mainstream methods for USI prediction into two categories: 

\begin{itemize}[leftmargin=*]
    \item \textbf{Task-specific supervised learning} commonly realizes USI prediction in a fully supervised task, e.g., identifying poverty levels~\cite{han2020learning,ayush2020generating,ayush2021efficient}, crop yields \cite{russwurm2020self,martinez2021fully,m2019semantic,yeh2021sustainbench}, and commercial activeness \cite{he2018perceiving,liu2023knowledge}. 
However, its inherent task specificity (\emph{i.e.,} reliant on ample labeled data), may impede the model's capacity for broad generalization to other downstream tasks. 

    \item \textbf{Vision pretraining}, also known as a type of Self-Supervised Learning~\cite{jaiswal2020survey}, aims to learn general visual features from urban satellite imagery, which are subsequently fine-tuned on a specific task for enhancing performance 
    ~\cite{BAI2023193,xi2022beyond,liu2023knowledge}. The inclusion of more auxiliary data, \emph{e.g.}, Points of Interests (POIs)~\cite{zhang2021multi,limcn4rec} and road networks~\cite{liang2018geoman,liang2021fine,chen2023tul},
    leads to richer, more accurate, and more useful representations of urban areas. For clarity, Figure~\ref{fig:intro} (a-b) depict a sketch of these two streams of approaches.
\end{itemize}

Despite the success of utilizing satellite imagery for USI prediction, urban environments exhibit a spatial hierarchy in reality, from a macro \emph{region} level to a micro \emph{location} level (\emph{e.g.}, architectural details, street furniture). Previous frameworks in Figure \ref{fig:intro} (a-b) mainly focus on single granularity, neglecting finer-grained visual clues.
Nonetheless, as the saying goes, \emph{``the devil is in the details.''}. By zooming into micro levels using corresponding street-view images, a more nuanced and fine-grained understanding emerges, as shown in the Figure~\ref{fig:intro}(d) Left. Therefore, the integration and alignment of multi-granularity information remain ongoing challenges in USI prediction that require further exploration.


In recent years, 
by leveraging the extensive knowledge embedded in LLMs~\cite{manvi2023geollm} and the inherent interpretability of text modality, a wide range of tasks across various domains are empowered.
Textual data can serve as an effective auxiliary tool in multimodal learning across a wide range of scenarios, such as geo-localization~\cite{ligeoreasoner} and recommender systems~\cite{gao2024llm}.

While in USI prediciton, the efficacy of leveraging textual modality for semantic enrichment is still less explored. UrbanCLIP~\cite{yan2023urban} stands out as the innovative solution, which generates text description by a pretrained LLM for satellite imagery and achieves learning urban region representations through natural language supervision.

Nevertheless, the text generation process of UrbanCLIP raises several critical concerns:
\emph{i) Hallucination}: The generation of textual content sometimes deviates from or introduces information not present in the input satellite imagery. 
\emph{ii) Homogenization}: The generated descriptions appear overly simplified and general, potentially leading to homogenization and inadequate differentiation. To better align with the intricacies of satellite (or street-view) imagery, \emph{we necessitate a more powerful approach that goes beyond the intuitive text generation method in UrbanCLIP, ensuring a more faithful and explainable representation of urban regions}.

In this paper, we present a \underline{V}ision-\underline{L}anguage \underline{P}retraining framework (i.e., \textbf{UrbanVLP}) for \underline{urban} socioeconomic indicator prediction. As depicted in Figure~\ref{fig:intro}(d), our model elaborately integrates \underline{multi-granularity information} from both satellite (macro-level) and street-view (micro-level) imagery to produce comprehensive urban region representations, while simultaneously harnessing the interpretability inherent in \underline{high-quality textual descriptions}. Targeting the first challenge, we introduce a novel \emph{Multi-Granularity Cross-Modal Alignment} module, which utilizes dual-branch contrastive learning to establish alignment between information derived from two semantic granularities. 
To address the second issue (\emph{i.e.}, hallucination and homogenization in LLM generated texts), we devise an \emph{Automatic Text Generation} together with a \emph{Calibration} mechanism to uphold text quality standards. 
To guarantee the quality of LLM-generated descriptions, we propose a reference-free metric called~\texttt{PerceptionScore}, motivated by the human evaluation system.

\begin{figure}
    \centering
\includegraphics[width=\columnwidth]{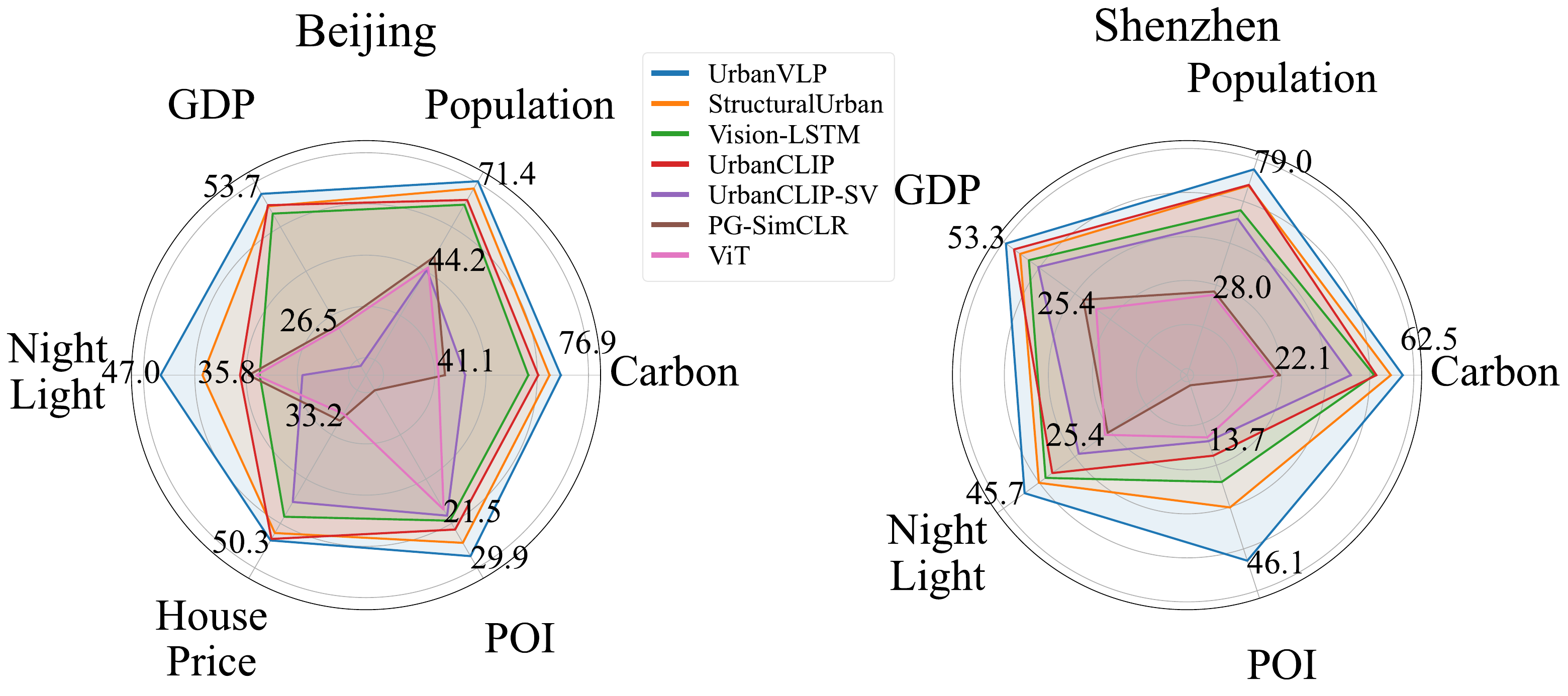}
    \caption{$R^2$ results in Beijing and Shenzhen.}
    \label{fig:radarimg}
    \vspace{-3mm}
\end{figure}

In summary, our contributions lie in the following aspects:
\vspace{-0.8em}

\begin{itemize}

\item \emph{Multi-Granularity Cross-Modal Alignment}. We explore the role of two distinct visual data modalities at various semantic granularities --  satellite imagery and street-view imagery. 
We inject fine-grained semantic information by integrating street-view data through token-level contrastive learning.

\vspace{-0.1em}
\item  \emph{Automatic Text Generation and Calibration}. Powered by image-to-text LLMs, we generate text descriptions and implement a robust evaluation mechanism based on a new reference-free metric, ensuring the fidelity between the generated texts and the corresponding image content. 

\item \emph{A New Benchmark \& Empirical Evidence}. 
We plan to open-source the first vision-text and multi-granularity urban dataset upon paper notification, consisting of six downstream tasks across the socio-economy. Extensive experiments
demonstrate that our UrbanVLP outperforms existing approaches by an average improvement of
3.95\% on the $R^2$ metric, as illustrated in Figure~\ref{fig:radarimg}.

\end{itemize}

\begin{figure*}[!t]
    \centering
    \includegraphics[width=1.0\linewidth]
    {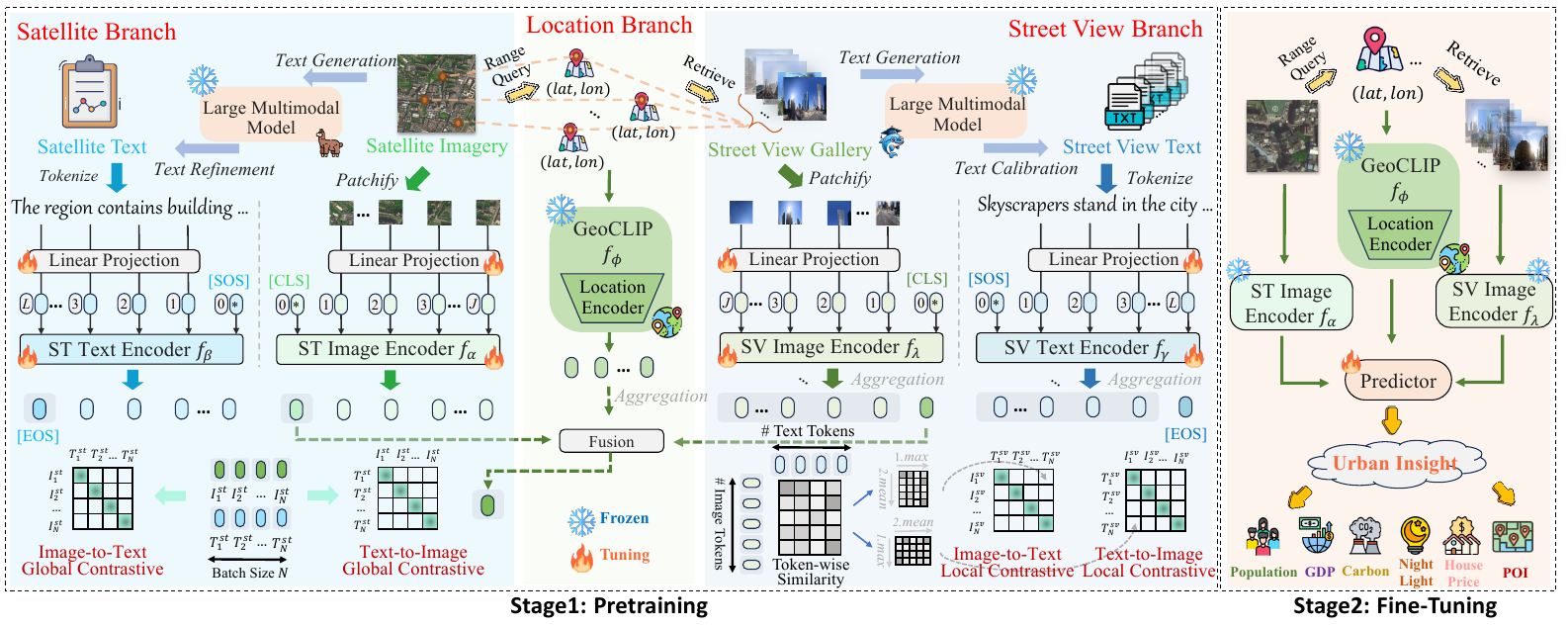} 
    \vspace{-2em}
    \caption{Overall framework of our proposed \model.}
    \label{fig:overall_framework_pretrain}
    \vspace{-1.5em}
\end{figure*}

\section{Related Work}

We formally define the problem of USI prediction. Given a satellite image $I^{sa}_g$, a set of street-view images $\mathcal{I}^{sv}_g$ belonging to the target coverage, and their corresponding latitude and longitude pair $\mathcal{L}_g$. The objective is to employ a learning function $\mathcal{F}$  to map them to representation vectors $\mathbf{e}_g=\mathcal{F}(I^{sa}_g, \mathcal{I}^{sv}_g, \mathcal{L}_g)$. The representation can then be further utilized in downstream tasks, such as inferring socioeconomic indicators $\mathbf{Y}_g$ for the given region $g$. 
The detailed formulation and definitions related to this task are provided in Appendix~A. 

\vspace{0.5em}

\noindent \textbf{Urban Socioeconomic Indicator Prediction.} 
In literature, many studies have focused on learning task-specific region representations from various urban data,
especially urban imagery due to its consistent updates and easy accessibility \cite{liu2023knowledge,xi2022beyond}.
For example, Urban2Vec \cite{wang2020urban2vec} integrated \emph{street-view imagery} and POI data to learn neighborhood embeddings. Some contrastive learning approaches like PG-SimCLR \cite{xi2022beyond} and UrbanCLIP \cite{yan2023urban} have shown success in representing urban regions through \emph{satellite images}.
However, the aforementioned works exclusively considered a single type of urban imagery, \emph{overlooking the potential synergy between satellite imagery and street-view imagery, which can complement each other.}
\vspace{0.1cm}

\noindent \textbf{Vision-Language Pretraining (VLP).} 
VLP aims to jointly encode vision and language in a fusion model.
A milestone work CLIP \cite{radford2021learning} and its variants \cite{li2021supervision,yao2022filip} highlight the efficacy of contrastive learning in cross-modal downstream tasks, such as zero-shot learning and cross-modal retrieval.
Recent works \cite{tsimpoukelli2021multimodal,alayrac2022flamingo} shift towards leveraging LLMs knowledge for VLP. \emph{In USI prediction, the potential of VLP paradigm remains untapped, with limited exploration of the benefits of textual information.}

\section{Methodology}

Figure \ref{fig:overall_framework_pretrain} shows our framework with two stages.
In the pretraining stage, we first devise an automatic text generation and calibration module using ShareGPT4V~\cite{chen2023sharegpt4v} to generate textual descriptions for street-view images with geographical and visual prompts. To guarantee the quality of the generated text, we introduced a novel metric called \texttt{PerceptionScore}.
Then a multi-granularity cross-modal alignment framework is designed to utilize multi-level contrastive learning and fine-grained information injection.
During the fine-tuning stage, we employ frozen encoders from Stage 1 to extract features and fine-tune a lightweight MLP for accurate predictions. 

\vspace{-0.5em}
\subsection{Automatic Text Generation and Calibration}
\label{sec:text generation}

\noindent \textbf{Text Generation.} 
For each street-view image, we use advanced Large Multimodal Models (LMMs) due to their robust cross-modal knowledge retention abilities, to provide comprehensive textual descriptions.
However, this confronts two primary challenges. Firstly, due to \textit{substantial financial costs} associated with API usage, employing closed source LMMs like GPT4V \cite{yang2023dawn} or Gemini \cite{fu2023challenger} for generating tens of thousands of text descriptions is impractical. Secondly, street-view images often encompass diverse details, \textit{necessitating a well-designed prompt template} for obtaining high-quality textual descriptions.

To address these issues, we first employ an open-source model closely related to GPT-4V, known as ShareGPT4V \cite{chen2023sharegpt4v}, to generate the street-view image descriptions. This model is a powerful LLM with vision capabilities trained on 100k GPT4V-generated captions, revealing captioning ability comparable to GPT4V.
Moreover, in the process of designing text prompt templates, we believe that the proportion of each element in street-view images reflects their relative abundance, thereby facilitating LMMs in evaluating their significance. 
Therefore, we employ a pretrained segmentation model~\cite{githubGitHubCSAILVisionsemanticsegmentationpytorch} to decouple visual elements and calculate the proportion of segmentation for each element. 
As evidenced by existing research~\cite{fan2023urban}, the accuracy of current segmentation methods is already adequate to recognize various visual components.
Simultaneously, geospatial coordinates of street-view images are incorporated as prompts to aid in the generation of more precise textual descriptions. 
Template and segmentation ratio details can be found in Appendix~H.

\noindent \textbf{Text Calibration.} 
Existing works~\cite{fan2023improving, yan2023urban} have already demonstrated that the quality of the textural descriptions is crucial for enhancing model performance. 
However, previous research~\cite{yan2023urban} primarily employs simplistic rules and manual refinement processes for refining text description of satellite imagery, leading to potential unresolved hallucination problems~\cite{rawte2023survey}. Besides, the rule-based methods are overly general and fail to address specific issues in each image description as they are tailored to particular cases. On the other hand, manual rewriting is highly labor-intensive, which limits its scalability.

\vspace{-0.15em}
Therefore, an effective method for assessing the quality of generated text and calibrating the alignment between the text and image is essential to ensure a high level of LMM-inherent knowledge integration. Compared to classical methods that assess LMM-generated descriptions by relying on manually-annotated references to measure content overlap~\cite{DBLP:conf/iclr/ZhangKWWA20, lee2020vilbertscore}, we devise an automatic mechanism to simulate human evaluation systems. 
This mechanism places greater emphasis on semantic consistency rather than merely assessing typical content overlap, by introducing a novel quality metric called \texttt{PerceptionScore}.
Concretely, it comprises two parts: text semantic quality and visual recall quality.
The former one directly adopts the efficient implementation of \texttt{CLIPScore}~\cite{hessel-etal-2021-clipscore}, which is based on CLIP's strong zero-shot capabilities and demonstrates high correlation with human semantic judgment. 
However, \texttt{CLIPScore} solely measures textual quality without consideration of the omission of specific visual elements in the text, thereby \textit{lacking an assessment of visual recall}~\cite{hu2023infometic}.

\begin{figure}
     \centering
\includegraphics[width=1.0\columnwidth]
    {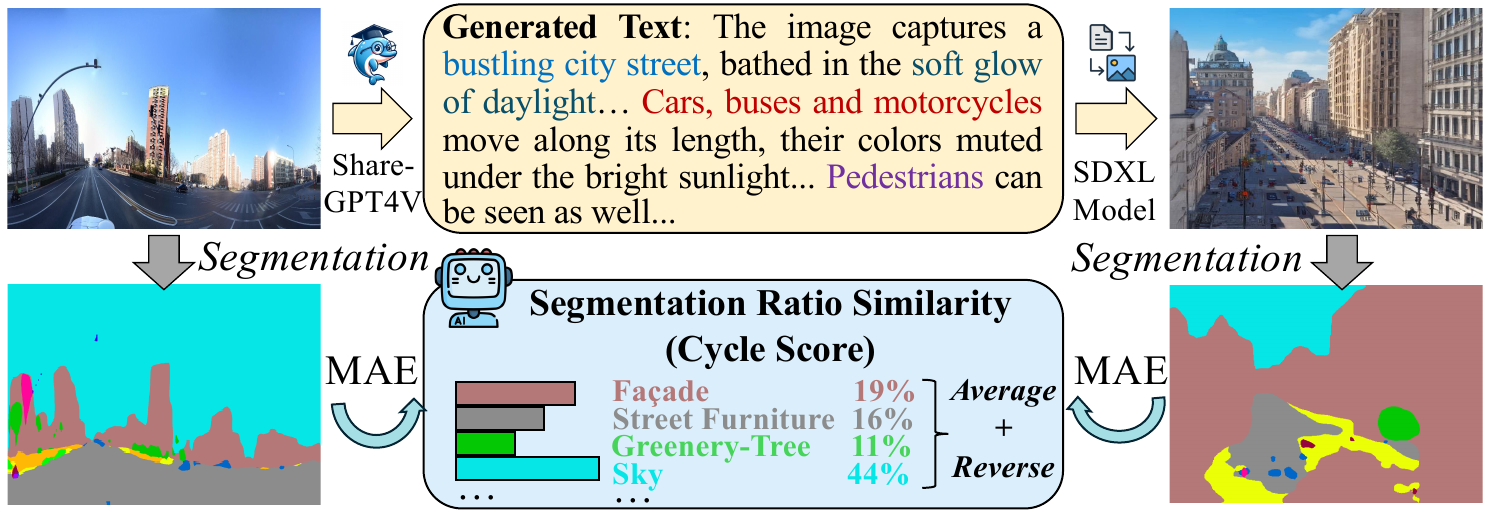}
    \vspace{-1.5em}
    \caption{The procedure of \texttt{CycleScore} calculation.}
    \label{fig:cyclescore}
    \vspace{-1.5em}
\end{figure}

To address this problem, we propose \texttt{CycleScore}, a metric designed to align fine-grained elemental information within the image $I$.
As illustrated in Figure \ref{fig:cyclescore}, given each description $T$ generated by LMMs, we leverage the state-of-the-art open-sourced text-to-image model $\operatorname{SDXL}(\cdot)$~\cite{podell2023sdxl} to generate a consistent image $I'$, reflecting various information contained in the text intuitively. Subsequently, we utilize a pretrained segmentation model 
$\operatorname{Seg}(\cdot)$~\cite{githubGitHubCSAILVisionsemanticsegmentationpytorch} to decouple consistent semantic elements~\cite{fan2023urban} in both the input and output images and calculate the Mean Absolute Error, \ie~$\operatorname{MAE}(\cdot)$ score, enforcing visual-semantic consistency. We formalize the process as follows:

\vspace{-1em}
\begin{small}
\begin{flalign}
&I' = \operatorname{SDXL}(I), &&\\
&\texttt{CycleScore}(I,I') = 1-\operatorname{MAE}(\operatorname{Seg}(I), \operatorname{Seg}(I')), &&\\
&\texttt{PerceptionScore}(I,T) = (\texttt{CLIPScore}(I,T) + && \nonumber \\
&\phantom{\texttt{PerceptionScore}(I,T)=(}\mathop{\texttt{CycleScore}}(I,I'))/2. &&
\end{flalign}
\end{small}

\vspace{-0.8em}
\subsection{Multi-Granularity Cross-modal Alignment}

\noindent \textbf{Modality Representation.} 
Utilizing automated text generation methods, we obtain a dataset of high-quality image-text pairs $D=(I, T)$. Here, $I$ represents satellite images $I^{st}$ or street-view images $I^{sv}$, while $T$ indicates satellite text descriptions $T^{st}$ or street-view text descriptions $T^{sv}$.

We first deploy Vision Transformer (ViT) \cite{dosovitskiy2021an} as the image encoder to process images derived from satellite and street-view sources.
Then we use the output $\mathbf{z}_{l}^{0}$ corresponding to the global tokens ${I}_{cls}$ inserted at the beginning as the final output of the image encoder, rewrite as $\mathbf{z}_I$ for image representation.
Concurrently, we leverage the basic Transformer-Encoder \cite{vaswani2017attention} as the text encoder to comprehend and encode the semantics of textual descriptions in parallel. 
Furthermore, considering that the geospatial location information associated with street-view images can assist in pinpointing the precise geospatial coordinates of the area, 
we utilize the open-sourced GeoCLIP's \cite{cepeda2023geoclip} location encoder to capture longitude and latitude $\mathcal{L}_g$, generating semantically informative geospatial features, denoted as $\mathbf{z}_L$.
Additional details can be found in Appendix~G.

\noindent \textbf{Modality Alignment.}
To capture the comprehensive visual information, beyond deriving global  visual features for the region from satellite images, the aggregation of location-level clues within the region can enhance the injection of details.
Therefore, we further incorporate fine-grained information from street-view images, along with the corresponding geographical features. 
The formalized process is as follows:
\begin{equation}
\small
\textstyle \mathbf{z}_{g_{st}} = \textit{f}~(\mathbf{z}_{I_{st}}, ~ \operatorname{Aggr}(\mathbf{z}_{I_{sv}}^{1},...,\mathbf{z}_{I_{sv}}^{m}), ~\operatorname{Aggr}(\mathbf{z}_{L}^{1},...,\mathbf{z}_{L}^{m})),
\end{equation}
where $\mathbf{z}_{I_{st}}$ denotes the satellite visual features for the query region, $\mathbf{z}_{I_{sv}}^{k}$ is the $k$-th street-view visual feature in that region, and $\mathbf{z}_{L}^{k}$ represents the $k$-th street-view location feature in the same region. $m$ indicates the count of street-view scenes within the region. $\operatorname{Aggr}(\cdot)$ denotes the aggregation method. 
The function $\textit{f}~$ signifies the feature fusion method.

To establish a global-level semantic correspondence between the satellite features and their associated textual features, we employ a joint contrastive optimization approach for the image and text encoders in the satellite branch. 
For the $i$-th satellite image-text pair $(I^{st}_{i}, T^{st}_{i})$ in a mini-batch, we contrast the image-text pairs against others within the samples, 
which aims to maximally preserve the mutual information between the pairs in latent space.
The global contrastive loss function \( \mathcal{L}_{CG} \) is composed of two terms: \( \mathcal{L}_{{CG}}^{Image \rightarrow Text} \) and \( \mathcal{L}_{{CG}}^{Text \rightarrow Image} \), which measures the similarity between the visual and textual embeddings, respectively, defined as:
\begin{equation}
\small
\begin{aligned}
    \mathcal{L}_{CG} = -\frac{1}{N} \Bigg(&\sum_i^N \log \frac{\exp(I_{g_i}^{\top}T_{g_i}/\tau)}{\sum_{j=0}^N \exp(I_{g_i}^{\top}T_{g_j}/\tau)} \\
    &+ \sum_i^N \log \frac{\exp(T_{g_i}^{\top}I_{g_i}/\tau)}{\sum_{j=0}^N \exp(T_{g_i}^{\top}I_{g_j}/\tau)}\Bigg),
\end{aligned}
\end{equation}
where $I_{g_i}$ and $T_{g_j}$ are the normalized embedding of satellite representation $\mathbf{z}_{g_{st}}$ in the $i$-th pair and that of textual representation $\mathbf{z}_{T_{st}}$ in the $j$-th pair, respectively. Besides, $N$ is the batch size, and $\tau$ is the temperature for contrastive learning.

The street-view branch aims to enrich region embedding with fine-grained local-level information. 
However, previous approaches like \cite{radford2021learning,jia2021scaling} rely exclusively on global feature similarity within each modality, disregarding the necessity for fine-grained alignment, such as the correspondence between visual objects and textual tokens \cite{yao2022filip}.
To address this challenge, we leverage a fine-grained interaction mechanism to implement cross-modal alignment at the local level. Specifically, we utilize token-level maximum similarity between visual and textual tokens to direct the contrastive objective.
We first compute the similarity between each visual token and all textual tokens, and then leverage the maximum value to calculate the average similarity of all image tokens to textual tokens. The similar approach is also applied to text-to-image process.
\begin{equation}
\small
\textstyle 
\operatorname{SIM}(v_i, t_i) = \frac{1}{l_1}(\sum_{k_1=1}^{l_1}{\operatorname{argmax}_{k_2\in[0,l_2)}(v_{ik_1}^{\top}t_{ik_{2}})}),
\end{equation}
\begin{equation}
\textstyle 
\operatorname{SIM}(t_i, v_i) = \frac{1}{l_2}(\sum_{k_2=1}^{l_1}{\operatorname{argmax}_{k_2\in[0,l_1)}(t_{ik_1}^{\top}v_{ik_{2}})}),
\end{equation}
where $v_i$ and $t_{j}$ denote the normalized embedding of street-view representation $\mathbf{z}_{I_{sv}}$ in the $i$-th pair and that of textual representation $\mathbf{z}_{T_{sv}}$ in the $j$-th pair, respectively. The fine-grained token-level representation can be optimized via:
\begin{equation}
\small
\begin{aligned}
    \mathcal{L}_{CL} = -\frac{1}{N} \Bigg(&\sum_i^N \log \frac{\exp(\operatorname{SIM}(v_{i}^{\top}, t_{i})/\tau)}{\sum_{j=0}^N \exp(\operatorname{SIM}(v_{i}^{\top}, t_{j})/\tau)} \\
    &+ \sum_i^N \log \frac{\exp(\operatorname{SIM}(t_{i}^{\top}, v_{i})/\tau)}{\sum_{j=0}^N \exp(\operatorname{SIM}(t_{i}^{\top}, v_{j})/\tau)}\Bigg).
\end{aligned}
\end{equation}

\begin{table*}[t!]
\setlength{\tabcolsep}{0.85mm}{}
\begin{small}
\scalebox{0.8}{
\begin{tabular}{c|c|ccc|ccc|ccc|ccc|ccc|ccc|ccc}

\toprule

\multicolumn{2}{c|}{Methods} & \multicolumn{3}{c|}{\model} &\multicolumn{3}{c|}{StructuralUrban} & \multicolumn{3}{c|}{Vision-LSTM} & \multicolumn{3}{c|}{UrbanCLIP-SV} & \multicolumn{3}{c|}{UrbanCLIP} & \multicolumn{3}{c|}{PG-SimCLR} & \multicolumn{3}{c}{ViT } \\

\midrule

\multicolumn{2}{c|}{Metric} & \cellcolor{blue!3}$\text{R}^2$ & \cellcolor{orange!3}RMSE & \cellcolor{green!3}MAE & \cellcolor{blue!3}$\text{R}^2$ & \cellcolor{orange!3}RMSE & \cellcolor{green!3}MAE & \cellcolor{blue!3}$\text{R}^2$ & \cellcolor{orange!3}RMSE & \cellcolor{green!3}MAE& \cellcolor{blue!3}$\text{R}^2$ & \cellcolor{orange!3}RMSE & \cellcolor{green!3}MAE & \cellcolor{blue!3}$\text{R}^2$ & \cellcolor{orange!3}RMSE & \cellcolor{green!3}MAE& \cellcolor{blue!3}$\text{R}^2$ & \cellcolor{orange!3}RMSE & \cellcolor{green!3}MAE & \cellcolor{blue!3}$\text{R}^2$ & \cellcolor{orange!3}RMSE & \cellcolor{green!3}MAE
\\

\midrule
\multirow{6}{*}{\rotatebox{90}{Beijing}}
& Carbon & \cellcolor{blue!3}\textbf{0.769} & \cellcolor{orange!3}\textbf{0.477} & \cellcolor{green!3}\textbf{0.369} & \cellcolor{blue!3}\underline{0.736} & \cellcolor{orange!3}\underline{0.518} & \cellcolor{green!3}\underline{0.405} & \cellcolor{blue!3}0.674 & \cellcolor{orange!3}0.519 & \cellcolor{green!3}0.433 & \cellcolor{blue!3}0.489 & \cellcolor{orange!3}0.713 & \cellcolor{green!3}0.548 & \cellcolor{blue!3}0.703  & \cellcolor{orange!3}0.541  & \cellcolor{green!3}0.539 & \cellcolor{blue!3}0.430 & \cellcolor{orange!3} 0.797  & \cellcolor{green!3}0.632 & \cellcolor{blue!3}0.411   & \cellcolor{orange!3}0.810  & \cellcolor{green!3}0.607 
\\

& Population & \cellcolor{blue!3}\textbf{0.714} & \cellcolor{orange!3}\textbf{0.523} & \cellcolor{green!3}\textbf{0.411} & \cellcolor{blue!3}\underline{0.691} & \cellcolor{orange!3} \underline{0.545}  & \cellcolor{green!3}\underline{0.427} & \cellcolor{blue!3}0.640  & \cellcolor{orange!3}0.591  & \cellcolor{green!3}0.518 & \cellcolor{blue!3}0.435  & \cellcolor{orange!3}0.734  & \cellcolor{green!3}0.581 & \cellcolor{blue!3}0.655  & \cellcolor{orange!3}0.576 & \cellcolor{green!3}0.459 & \cellcolor{blue!3}0.476  & \cellcolor{orange!3}1.228  & \cellcolor{green!3}0.963 & \cellcolor{blue!3}0.442  & \cellcolor{orange!3}0.861  & \cellcolor{green!3}0.635 
\\

& GDP & \cellcolor{blue!3}\textbf{0.537} & \cellcolor{orange!3}\textbf{0.684} & \cellcolor{green!3}\textbf{0.416} & \cellcolor{blue!3}0.512  & \cellcolor{orange!3}\underline{0.694}  & \cellcolor{green!3}\underline{0.426} & \cellcolor{blue!3}0.497  & \cellcolor{orange!3}0.717  & \cellcolor{green!3}0.469 & \cellcolor{blue!3}0.188  & \cellcolor{orange!3}0.910  & \cellcolor{green!3}0.568 &  \cellcolor{blue!3}\underline{0.514} & \cellcolor{orange!3} 0.694 & \cellcolor{green!3}0.445 & 
\cellcolor{blue!3}0.270  & \cellcolor{orange!3}1.679  & \cellcolor{green!3}1.067 &
\cellcolor{blue!3}0.265  & \cellcolor{orange!3}1.073  & \cellcolor{green!3}0.730 
\\

& Night Light & \cellcolor{blue!3}\textbf{0.470} & \cellcolor{orange!3}\textbf{0.668} & \cellcolor{green!3}\textbf{0.403} & \cellcolor{blue!3}\underline{0.421}  & \cellcolor{orange!3}\underline{0.696}  & \cellcolor{green!3}0.459 & \cellcolor{blue!3}0.354  & \cellcolor{orange!3}0.747  & \cellcolor{green!3}0.475 & \cellcolor{blue!3}0.304  & \cellcolor{orange!3}0.769  & \cellcolor{green!3}0.483 & \cellcolor{blue!3}0.377 & \cellcolor{orange!3} 0.741 & \cellcolor{green!3}0.468  & 
\cellcolor{blue!3}0.367 & \cellcolor{orange!3}0.728  & \cellcolor{green!3}\underline{0.404} & 
\cellcolor{blue!3}0.358  & \cellcolor{orange!3}0.733  & \cellcolor{green!3}0.523 
\\

& House Price & \cellcolor{blue!3}\textbf{0.503} & \cellcolor{orange!3}\textbf{0.644} & \cellcolor{green!3}\textbf{0.482} & \cellcolor{blue!3}0.493  & \cellcolor{orange!3}0.649  & \cellcolor{green!3}\underline{0.485} & \cellcolor{blue!3}0.471  & \cellcolor{orange!3}0.658 & \cellcolor{green!3} 0.495  & \cellcolor{blue!3}0.451  & \cellcolor{orange!3}0.674  & \cellcolor{green!3}0.515 & \cellcolor{blue!3}\underline{0.501}  & \cellcolor{orange!3}\underline{0.647}  & \cellcolor{green!3}0.486  & \cellcolor{blue!3}0.341  & \cellcolor{orange!3}0.718  & \cellcolor{green!3}0.555 & \cellcolor{blue!3}0.332  & \cellcolor{orange!3}0.719  & \cellcolor{green!3}0.569   
\\

& POI & \cellcolor{blue!3}\textbf{0.299} & \cellcolor{orange!3}\textbf{0.723} & \cellcolor{green!3}\textbf{0.374} & \cellcolor{blue!3}\underline{0.275} & \cellcolor{orange!3}\underline{0.725}  & \cellcolor{green!3}\underline{0.380} & \cellcolor{blue!3}0.235  & \cellcolor{orange!3}0.728  & \cellcolor{green!3}0.388 & \cellcolor{blue!3}0.226  & \cellcolor{orange!3}0.729  & \cellcolor{green!3}0.407 & \cellcolor{blue!3}0.251  & \cellcolor{orange!3}0.734  & \cellcolor{green!3}0.385 & \cellcolor{blue!3}- & \cellcolor{orange!3}- & \cellcolor{green!3}- & \cellcolor{blue!3}0.215 & \cellcolor{orange!3}0.742  & \cellcolor{green!3}0.411 
\\

\midrule

\multirow{6}{*}{\rotatebox{90}{Shanghai}}
& Carbon & \cellcolor{blue!3}\textbf{0.718} & \cellcolor{orange!3}\underline{0.520} & \cellcolor{green!3}\textbf{0.381} & \cellcolor{blue!3}\underline{0.678}  & \cellcolor{orange!3}0.550  & \cellcolor{green!3}0.432 & \cellcolor{blue!3}0.571  & \cellcolor{orange!3}0.606  & \cellcolor{green!3}0.455 & \cellcolor{blue!3}0.576  & \cellcolor{orange!3}\textbf{0.465}  & \cellcolor{green!3}0.450 & \cellcolor{blue!3}0.673  & \cellcolor{orange!3}0.560  & \cellcolor{green!3}\underline{0.426} & \cellcolor{blue!3}0.270  & \cellcolor{orange!3}0.742  & \cellcolor{green!3}0.551 & \cellcolor{blue!3}0.254   & \cellcolor{orange!3}0.758   & \cellcolor{green!3}0.522 
\\

& Population & \cellcolor{blue!3}\textbf{0.589} & \cellcolor{orange!3}\textbf{0.609} & \cellcolor{green!3}\textbf{0.476} & \cellcolor{blue!3}\underline{0.545}  & \cellcolor{orange!3}\underline{0.631} & \cellcolor{green!3}\underline{0.488} & \cellcolor{blue!3}0.518 & \cellcolor{orange!3}0.659 & \cellcolor{green!3}0.522  & \cellcolor{blue!3}0.400  & \cellcolor{orange!3}0.793  & \cellcolor{green!3}0.592 & \cellcolor{blue!3}0.533  & \cellcolor{orange!3}0.650  & \cellcolor{green!3}0.511 & \cellcolor{blue!3}0.288  & \cellcolor{orange!3}1.051  & \cellcolor{green!3}0.831 & \cellcolor{blue!3}0.279 & \cellcolor{orange!3}0.826  & \cellcolor{green!3}0.627 
\\

& GDP & \cellcolor{blue!3}0.323 & \cellcolor{orange!3}\underline{0.805} & \cellcolor{green!3}\underline{0.593} & \cellcolor{blue!3}0.312 & \cellcolor{orange!3}0.840  & \cellcolor{green!3}0.613 & \cellcolor{blue!3}0.311 & \cellcolor{orange!3}1.008 & \cellcolor{green!3}0.661 & \cellcolor{blue!3}\underline{0.323} & \cellcolor{orange!3}1.214  & \cellcolor{green!3}0.741 & \cellcolor{blue!3}\textbf{0.334}  & \cellcolor{orange!3}\textbf{0.804}  & \cellcolor{green!3}\textbf{0.586} & \cellcolor{blue!3}0.270  & \cellcolor{orange!3}1.679   & \cellcolor{green!3}1.067 & \cellcolor{blue!3}0.263 & \cellcolor{orange!3}1.221  & \cellcolor{green!3}0.735 
\\

& Night Light & \cellcolor{blue!3}\textbf{0.435} & \cellcolor{orange!3}\textbf{0.683} & \cellcolor{green!3}\textbf{0.490} & \cellcolor{blue!3}\underline{0.411}  & \cellcolor{orange!3}\underline{0.690}  & \cellcolor{green!3}0.497 & \cellcolor{blue!3}0.352  & \cellcolor{orange!3}0.734  & \cellcolor{green!3}0.550 & \cellcolor{blue!3}0.274  & \cellcolor{orange!3}0.765  & \cellcolor{green!3}0.556 & \cellcolor{blue!3}0.384  & \cellcolor{orange!3}0.710  & \cellcolor{green!3}0.515 & \cellcolor{blue!3}0.228  & \cellcolor{orange!3}0.756  & \cellcolor{green!3}\underline{0.495} & \cellcolor{blue!3}0.209 & \cellcolor{orange!3}0.768  & \cellcolor{green!3}0.561 
\\

& POI & \cellcolor{blue!3}\textbf{0.381} & \cellcolor{orange!3}\textbf{0.680} & \cellcolor{green!3}\textbf{0.386} & \cellcolor{blue!3}\underline{0.375}  & \cellcolor{orange!3}\underline{0.685}  & \cellcolor{green!3}\underline{0.392} & \cellcolor{blue!3}0.246  & \cellcolor{orange!3}0.704  & \cellcolor{green!3}0.390 & \cellcolor{blue!3}0.240  & \cellcolor{orange!3}0.711  & \cellcolor{green!3}0.402 & \cellcolor{blue!3}0.265 & \cellcolor{orange!3}0.710  & \cellcolor{green!3}0.395 & \cellcolor{blue!3}- & \cellcolor{orange!3}- & \cellcolor{green!3}- & \cellcolor{blue!3}0.234 & \cellcolor{orange!3} 0.713  & \cellcolor{green!3}0.406 
\\

\midrule

\multirow{6}{*}{\rotatebox{90}{Guangzhou}}
& Carbon & \cellcolor{blue!3}\textbf{0.701} & \cellcolor{orange!3}\textbf{0.513} & \cellcolor{green!3}\textbf{0.372} & \cellcolor{blue!3}\underline{0.655} & \cellcolor{orange!3}0.550  & \cellcolor{green!3}0.504 & \cellcolor{blue!3}0.554  & \cellcolor{orange!3}0.564  & \cellcolor{green!3}0.457 & \cellcolor{blue!3}0.425  & \cellcolor{orange!3}\underline{0.520}  & \cellcolor{green!3}0.473  & \cellcolor{blue!3}0.585  & \cellcolor{orange!3}0.603  & \cellcolor{green!3}\underline{0.440} & \cellcolor{blue!3}0.413  & \cellcolor{orange!3}0.951  & \cellcolor{green!3}0.716 & \cellcolor{blue!3}0.374   & \cellcolor{orange!3}0.665   & \cellcolor{green!3}0.528
\\

& Population & \cellcolor{blue!3}\textbf{0.655} & \cellcolor{orange!3}\textbf{0.573} & \cellcolor{green!3}\textbf{0.454} & \cellcolor{blue!3}\underline{0.631} & \cellcolor{orange!3}\underline{0.587}  & \cellcolor{green!3}\underline{0.459} & \cellcolor{blue!3}0.614  & \cellcolor{orange!3}0.619  & \cellcolor{green!3}0.509 & \cellcolor{blue!3}0.491  & \cellcolor{orange!3}0.693  & \cellcolor{green!3}0.622 & \cellcolor{blue!3}0.627  & \cellcolor{orange!3}0.596  & \cellcolor{green!3}0.473 & \cellcolor{blue!3}0.294  & \cellcolor{orange!3}1.046  & \cellcolor{green!3}2.112 & \cellcolor{blue!3}0.263   & \cellcolor{orange!3}0.821   & \cellcolor{green!3}0.661 
\\

& GDP & \cellcolor{blue!3}\underline{0.442} & \cellcolor{orange!3}\underline{0.755} & \cellcolor{green!3}\textbf{0.541} & \cellcolor{blue!3}0.429  & \cellcolor{orange!3}0.781  & \cellcolor{green!3}0.549  & \cellcolor{blue!3}0.412  & \cellcolor{orange!3}0.997  & \cellcolor{green!3}0.624 & \cellcolor{blue!3}0.327  & \cellcolor{orange!3}0.998  & \cellcolor{green!3}0.690 & \cellcolor{blue!3}\textbf{0.446}  & \cellcolor{orange!3}\textbf{0.752}  & \cellcolor{green!3}\underline{0.546} & \cellcolor{blue!3}0.263  & \cellcolor{orange!3}1.157  & \cellcolor{green!3}1.553 & \cellcolor{blue!3}0.249   & \cellcolor{orange!3}1.122   & \cellcolor{green!3}0.719
\\

& Night Light & \cellcolor{blue!3}\textbf{0.548} & \cellcolor{orange!3}\textbf{0.594} & \cellcolor{green!3}\textbf{0.428} & \cellcolor{blue!3}\underline{0.508}  & \cellcolor{orange!3}\underline{0.607} & \cellcolor{green!3}0.448 & \cellcolor{blue!3}0.381  & \cellcolor{orange!3}1.021  & \cellcolor{green!3}0.629  & \cellcolor{blue!3}0.314  & \cellcolor{orange!3}0.692  & \cellcolor{green!3}0.524 & \cellcolor{blue!3}0.463  & \cellcolor{orange!3}0.651  & \cellcolor{green!3}0.483 & \cellcolor{blue!3}0.436  & \cellcolor{orange!3}0.649  & \cellcolor{green!3}\underline{0.432} & \cellcolor{blue!3}0.357  & \cellcolor{orange!3}0.642  & \cellcolor{green!3}0.505 
\\

& POI & \cellcolor{blue!3}\textbf{0.438} & \cellcolor{orange!3}\textbf{0.732} & \cellcolor{green!3}\textbf{0.333} & \cellcolor{blue!3}\underline{0.357} & \cellcolor{orange!3}\underline{0.758} & \cellcolor{green!3}\underline{0.359} & \cellcolor{blue!3}0.175  & \cellcolor{orange!3}0.811  & \cellcolor{green!3}0.401 & \cellcolor{blue!3}0.169 & \cellcolor{orange!3}0.826  & \cellcolor{green!3}0.405 & \cellcolor{blue!3}0.185  & \cellcolor{orange!3}0.802 & \cellcolor{green!3}0.401 & \cellcolor{blue!3}- & \cellcolor{orange!3}- & \cellcolor{green!3}- & \cellcolor{blue!3}0.136  & \cellcolor{orange!3}0.904  & \cellcolor{green!3}0.427 
\\

\midrule

\multirow{6}{*}{\rotatebox{90}{Shenzhen}}
& Carbon & \cellcolor{blue!3}\textbf{0.625} & \cellcolor{orange!3}\underline{0.605} & \cellcolor{green!3}\textbf{0.464} & \cellcolor{blue!3}\underline{0.587} & \cellcolor{orange!3}\textbf{0.593}  & \cellcolor{green!3}\underline{0.481} & \cellcolor{blue!3}0.534  & \cellcolor{orange!3}0.609  & \cellcolor{green!3}0.516 & \cellcolor{blue!3}0.460  & \cellcolor{orange!3}0.621  & \cellcolor{green!3}0.529  & \cellcolor{blue!3}0.541  & \cellcolor{orange!3}0.607  & \cellcolor{green!3}0.526 & \cellcolor{blue!3}0.234  & \cellcolor{orange!3}0.975  & \cellcolor{green!3}0.746 & \cellcolor{blue!3}0.221  & \cellcolor{orange!3}0.727  & \cellcolor{green!3}0.574  
\\

& Population & \cellcolor{blue!3}\textbf{0.790} & \cellcolor{orange!3}\textbf{0.452} & \cellcolor{green!3}\textbf{0.348} & \cellcolor{blue!3}0.724  & \cellcolor{orange!3}\underline{0.476}  & \cellcolor{green!3}\underline{0.386} & \cellcolor{blue!3}0.624  & \cellcolor{orange!3}0.562  & \cellcolor{green!3}0.471 & \cellcolor{blue!3}0.589  & \cellcolor{orange!3}0.604  & \cellcolor{green!3}0.484 & \cellcolor{blue!3}\underline{0.727}  & \cellcolor{orange!3}0.510  & \cellcolor{green!3}0.392 & \cellcolor{blue!3}0.294   & \cellcolor{orange!3}1.046   & \cellcolor{green!3}2.112 & \cellcolor{blue!3}0.280 & \cellcolor{orange!3}0.832 & \cellcolor{green!3}0.654 
\\

& GDP & \cellcolor{blue!3}\textbf{0.533} & \cellcolor{orange!3}\textbf{0.682} & \cellcolor{green!3}\textbf{0.447} & \cellcolor{blue!3}0.489 & \cellcolor{orange!3}0.697  & \cellcolor{green!3}0.471 & \cellcolor{blue!3}0.462  & \cellcolor{orange!3}0.723  & \cellcolor{green!3}0.517 & \cellcolor{blue!3}0.433  & \cellcolor{orange!3}0.755  & \cellcolor{green!3}0.577 & \cellcolor{blue!3}\underline{0.508}  & \cellcolor{orange!3}\underline{0.693}  & \cellcolor{green!3}\underline{0.464} & \cellcolor{blue!3}0.294  & \cellcolor{orange!3}1.046  & \cellcolor{green!3}2.112 & \cellcolor{blue!3}0.254  & \cellcolor{orange!3}1.226  & \cellcolor{green!3}0.779 
\\

& Night Light & \cellcolor{blue!3}\textbf{0.457} & \cellcolor{orange!3}\textbf{0.667} & \cellcolor{green!3}\textbf{0.459} & \cellcolor{blue!3}\underline{0.421}  & \cellcolor{orange!3}\underline{0.694} & \cellcolor{green!3}\underline{0.502} & \cellcolor{blue!3}0.404  & \cellcolor{orange!3}0.702  & \cellcolor{green!3}0.508 & \cellcolor{blue!3}0.320 & \cellcolor{orange!3}0.717  & \cellcolor{green!3}0.531 & \cellcolor{blue!3}0.387  & \cellcolor{orange!3}0.709  & \cellcolor{green!3}0.511  & \cellcolor{blue!3}0.247  & \cellcolor{orange!3}0.934  & \cellcolor{green!3}0.738 & \cellcolor{blue!3}0.254  & \cellcolor{orange!3}0.715  & \cellcolor{green!3}0.530  
\\

& POI & \cellcolor{blue!3}\textbf{0.461} & \cellcolor{orange!3}\textbf{0.752} & \cellcolor{green!3}\textbf{0.370} & \cellcolor{blue!3}\underline{0.321}  & \cellcolor{orange!3}\underline{0.780}  & \cellcolor{green!3}\underline{0.389} & \cellcolor{blue!3}0.254  & \cellcolor{orange!3}0.795  & \cellcolor{green!3}0.412 & \cellcolor{blue!3}0.147 & \cellcolor{orange!3}0.849  & \cellcolor{green!3}0.453 & \cellcolor{blue!3}0.185  & \cellcolor{orange!3}0.838 & \cellcolor{green!3}0.436 & \cellcolor{blue!3}- & \cellcolor{orange!3}- & \cellcolor{green!3}- & \cellcolor{blue!3}0.137 & \cellcolor{orange!3}0.954  & \cellcolor{green!3}0.455 
\\
\midrule

\multicolumn{2}{c|}{$1^{\text{st}}$Count} & \multicolumn{3}{c|}{56} & \multicolumn{3}{c|}{1} & \multicolumn{3}{c|}{0} & \multicolumn{3}{c|}{1} & \multicolumn{3}{c|}{5} & \multicolumn{3}{c|}{0} & \multicolumn{3}{c}{0}\\
\bottomrule
\end{tabular}
}
\vspace{-2mm}
\caption{Socioeconomic indicators prediction results in four datasets. The best results are in bold and the second-best results are \underline{underlined}.}
\label{tab:results_overall}
\vspace{-2em}
\end{small}
\end{table*}

\vspace{-0.5em}
\subsection{Pretraining \& Fine-Tuning}
\vspace{-0.2em}

\noindent \textbf{Pretraining Stage.} 
The overall objective of \model can be defined as the joint optimization of the above two losses:
\begin{equation}
\small
\mathcal{L}_{Total} = \alpha\mathcal{L}_{{CG}} + \beta\mathcal{L}_{{CL}}.
\label{eqa:hyperparam}
\end{equation}
where $\alpha$ and $\beta$ are hyperparameters for a trade-off. Through backpropagation optimization, we achieve multi-granularity cross-modal alignment, resulting in robust encoders.

\vspace{0.2em}
\noindent \textbf{Fine-Tuning Stage.} 
As depicted in Figure~\ref{fig:overall_framework_pretrain} Right, 
during the fine-tuning stage, we employ a linear probing approach \cite{he2022masked} which begins by extracting $\mathbf{e}_{st}, \mathbf{e}_{sv}, \mathbf{e}_{p}$ features from the pretrained encoder for satellite images, street-view images, and street-view positions. Subsequently, these features are fused together, and a minimalist classifier (MLP) is trained on top to fine-tune the prediction of urban metrics, denoted as $\mathbf{Y}_i = MLP(\mathbf{e}_{st}, \mathbf{e}_{sv}, \mathbf{e}_{p})$.
It is noteworthy that in downstream tasks, textual information is unnecessary since the knowledge embedded in the text has already been imparted to other modality encoders through pretraining, and additional text generation time hinders real-time application.

\vspace{-0.5em}
\section{Experiments}
\subsection{Experimental Setup}
\textbf{Dataset \& Task Description.} 
Given the current lack of open-sourced datasets in the research community, we introduce a new benchmark dataset named $\mathtt{\dataset}$, which will be released upon paper notification. $\mathtt{\dataset}$ is unique in that it comprises a dual-category structure encompassing both satellite and street-view image components, each paired with corresponding high-quality textual descriptions. 
We adhere to~\cite{yan2023urban} to cover core area data of four cities in China. A comprehensive overview of $\mathtt{\dataset}$, along with relevant statistics, is provided in Appendix~C. 

\vspace{0.2em}
\noindent \textbf{Baselines.}
Following the established practice~\cite{yan2023urban,xi2022beyond,liu2023knowledge} in this area, we compare our method with seven recent baselines in the field of imagery-based USI prediction. The single-granularity urban imagery-based methods include: 

\textbf{ViT}~\cite{dosovitskiy2021an},
\textbf{PG-SimCLR}~\cite{xi2022beyond}, \textbf{UrbanCLIP}~\cite{yan2023urban}, \textbf{UrbanCLIP-SV}~\cite{yan2023urban}.

We also apply multiple baselines for multi-granularity urban imagery-based USI prediction: \textbf{Vision-LSTM }\cite{HUANG2023102043} and \textbf{StructuralUrban}~\cite{li2022predicting}.
They are all trained using data from our $\mathtt{\dataset}$ dataset.
The in-depth overview of baselines is presented in Appendix~E.

\vspace{0.2em}
\noindent \textbf{Evaluation Metrics.}
We follow \cite{yan2023urban} to evaluate our method in terms of: the coefficient of determination $R^2$, the root mean squared error (RMSE), and the mean absolute error (MAE). 
An increase in $R^2$, along with a decrease in RMSE and MAE values, signifies improved model accuracy.

\vspace{0.2em}
\noindent \textbf{Implementation Details.} The implementation details are provided in Appendix~F.

\vspace{-0.5em}
\subsection{Performance Evaluation}
\label{sec:performance}
\vspace{-0.3em}

To evaluate our \model framework, we conduct comparisons with existing state-of-the-art methods on our proposed $\mathtt{\dataset}$ datasets. Table \ref{tab:results_overall} presents the overall results, from which we can obtain the following findings:

\vspace{0.2em}
\noindent \textbf{1) \model significantly outperforms the baselines which employ single/multi-granularity imagery.}
It can be seen that UrbanVLP surpasses the best baseline (StructuralUrban) by 3.3\%, 2.3\%, 2.5\%, 4.9\%, 1.0\% and 2.4\% in terms of $R^2$ for all six indicators: Carbon, Population, GDP, Night Light, House Price and POI in Beijing.
Similarly, we can witness consistent improvements in the other three cities.
In addition, the average reduction of \model in terms of RMSE and MAE on the Beijing dataset are 1.8\% and 2.1\%, respectively. 
For a more intuitive display, we visualize the overall performance on Beijing and Shenzhen in Figure \ref{fig:radarimg}, where the results further prove the versatility of our framework for USI prediction.

\vspace{0.2em}
\noindent \noindent \textbf{2) Integrating multi-granularity imagery can lead to superior performance against baselines.} This merit can be rationalized by the incorporation of additional fine-grained information derived from street-view imagery.
Meanwhile, UrbanCLIP-SV underperforms UrbanCLIP, revealing that street-view imagery (while rich in details) lacks a macro view, resulting in suboptimal outcomes when used in isolation. This also underscores the irreplaceability of satellite imagery serving as a macro visual modality.

\vspace{0.2em}
\noindent  \noindent \textbf{3) Compared to other modalities (e.g., POIs), the textual modality facilitates a more comprehensive understanding of the region.}
We include various baselines for comparison, such as PGSimCLR which incorporates POI distributions to integrate external information and enhance regional representations. Despite their promising results, our UrbanVLP illustrates the efficacy of integrating textual data as a information-compact modality~\cite{he2022masked}, which fully leverages the inherent knowledge of LLMs and explicitly enhances interpretability in the training phase.

\vspace{-0.7em}
\subsection{Ablation Studies}

\begin{figure}[!h]
    \centering
\includegraphics[width=1.0\columnwidth]{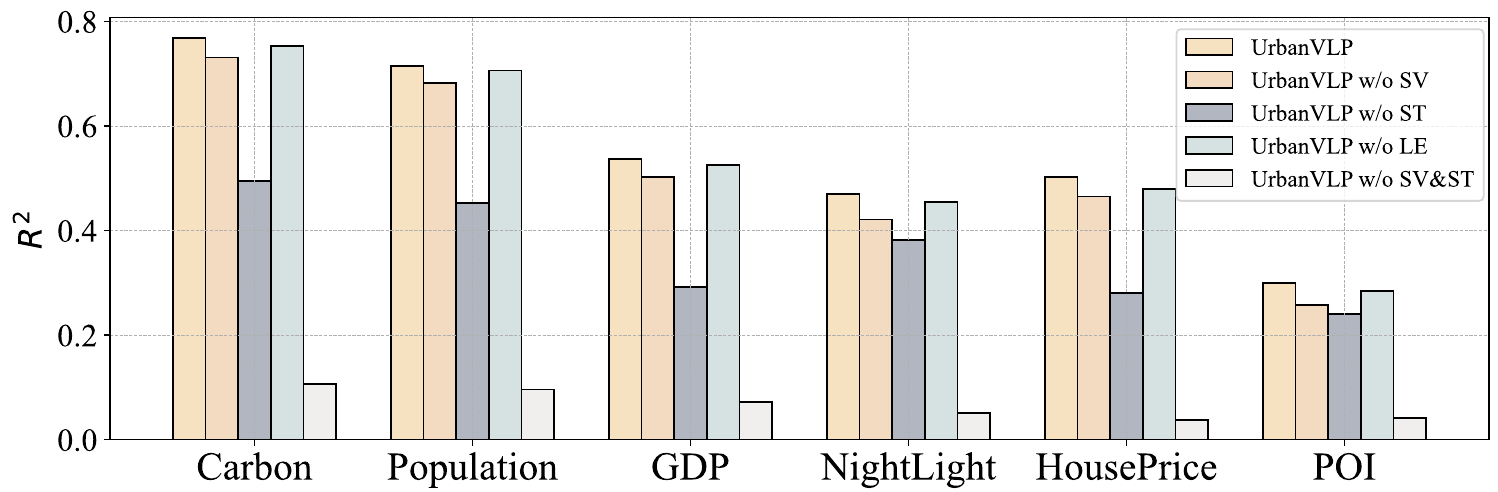}
    \vspace{-2em}
    \caption{Ablation study on $\mathtt{\dataset}$-$\mathtt{Beijing}$.}
    \label{fig:ablation}
    \vspace{-1em}
\end{figure}

As shown in Figure~\ref{fig:ablation}, we conduct ablation studies to examine each component in \model on the $\mathtt{\dataset}$-Beijing dataset, including the Satellite branch (ST branch), Street-View branch (SV branch), and Location Encoding branch (LE branch).
More discussion can be found in Appendix~J.

\noindent \textbf{Effects of Street-View Branch.} 
One of our contributions is the incorporation of multi-granularity information that reflects the urban spatial hierarchy. 
It can be observed that the incorporation of street-view branch results in an average improvement of 3.56\% in terms of $R^2$. This enhancement is attributed to the street-view branch's ability to capture rich, detailed information, thereby improving modeling precision.

\vspace{0.2em}
\noindent \textbf{Effects of Textual Modality.} 
In Table \ref{tab:results_overall}, we demonstrate the function of the textual modality by comparing UrbanVLP with a standard ViT-based model, which shares the same configuration as the unimodal visual encoder of UrbanVLP. We then utilize the visual features extracted without textual augmentation to predict downstream USI.
As we can see, the lack of textual information leads to substantial performance degradation, underscoring the critical role of textual modalities in attaining a comprehensive visual representation. Similar findings were also reported in~\cite{yan2023urban}.

\vspace{0.2em}
\noindent \textbf{Effects of Location Coordinates.} 
In Figure~\ref{fig:ablation}, we evaluate the performance of the model without the Location Encoding branch.
Although not as impactful as visual modalities, the inclusion of geospatial locations has yielded an average improvement of 1.5\% in $R^2$. 
It is noteworthy that we utilize a well-pretrained location encoder and freeze it within our framework. Consequently, its generalizability is ensured, preventing overfitting to specific cities in $\mathtt{\dataset}$ dataset.

\subsection{Qualitative Analysis}
We further investigate the quality of generated texts and the predictive performance of UrbanVLP in practice. \textit{More empirical analysis can be found in Appendix~J}.

\begin{figure}[!h]
    \centering
\includegraphics[width=\columnwidth]{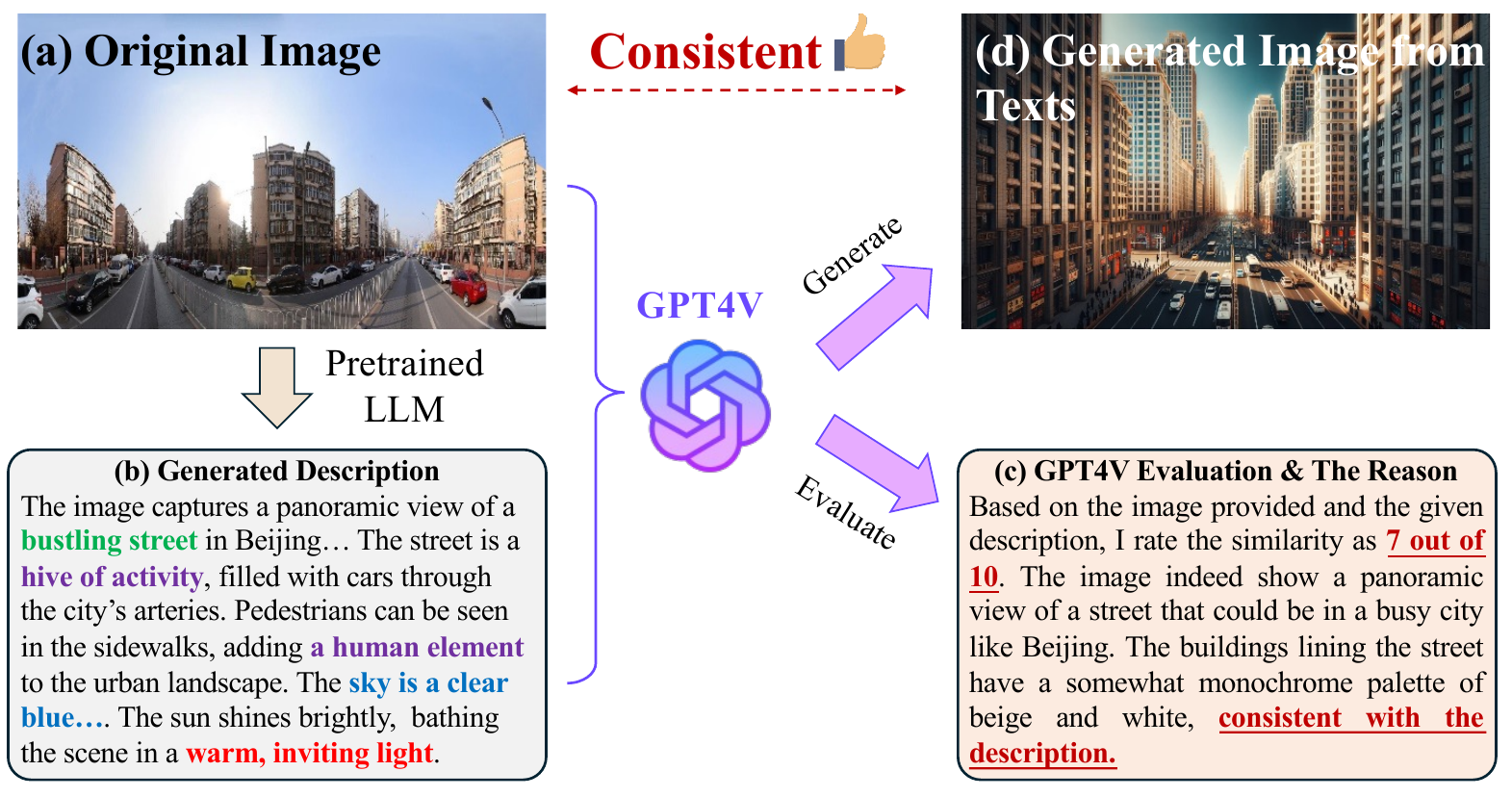}
\vspace{-0.5em}
    \caption{An example of generated text descriptions.}
    \label{fig:quality1}
\end{figure}

\vspace{0.2em}
\noindent \textbf{Illustration of the quality of our generated descriptions.} 
To visually illustrate the quality of the generated description, we present an example in Figure \ref{fig:quality1}. The generated description of the street-view image (a) is depicted in (b). Subsequently, we employ GPT4V \cite{yang2023dawn} to assess the quality score of the generated description, which is rated as 7 out of 10, indicating the effectiveness of the generated text. Furthermore, GPT4V also provides specific areas for improvement in (c). In (d), we utilize GPT4V to generate an image based on the description in (b). It is evident that the generated image bears a resemblance to the original one.

\begin{figure}[!h]
    \centering
\includegraphics[width=\columnwidth]{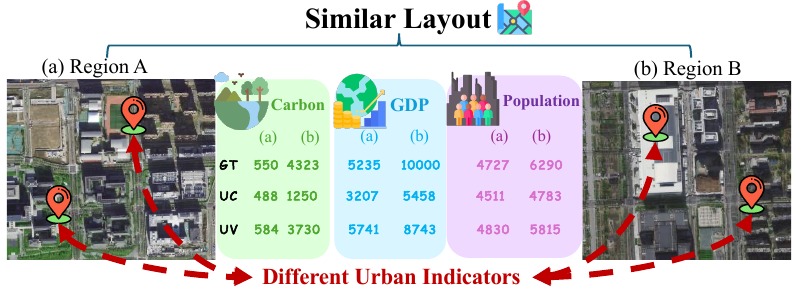}
\vspace{-0.5em}
    \caption{Case study of UrbanCLIP (UC) and our UrbanVLP (UV).}
    \label{fig:quality2}
\end{figure}

\noindent \textbf{Case Study for Predicted Results.} 
Here we also show some predicted results in Figure \ref{fig:quality2} on $\mathtt{\dataset}$-$\mathtt{Beijing}$ dataset. Satellite images (a) and (b), though similar in layout, differ significantly in land use. 
(a) encompasses residential and campus areas, whereas (b) represents industrial zones, leading to entirely different socioeconomic characteristics.
The carbon emissions and GDP of campus and residential area are significantly lower than those of industrial parks, whereas the disparities in population are not as pronounced, possibly due to the unique characteristics of school district housing near educational institutions.

As observed, UrbanCLIP's predictions fail to effectively differentiate between the two, as the downstream metrics in Figure~\ref{fig:quality2}(b) still lean towards predicting a homogenized socioeconomic area similar to (a). In contrast, the results from \model are capable of distinctly distinguishing the socioeconomic attributes of the two in terms of carbon emission and GDP.

\vspace{-0.5em}
\subsection{Visualization of Region Representations} 
\begin{figure}[h]
    \centering
\includegraphics[width=0.9\columnwidth]{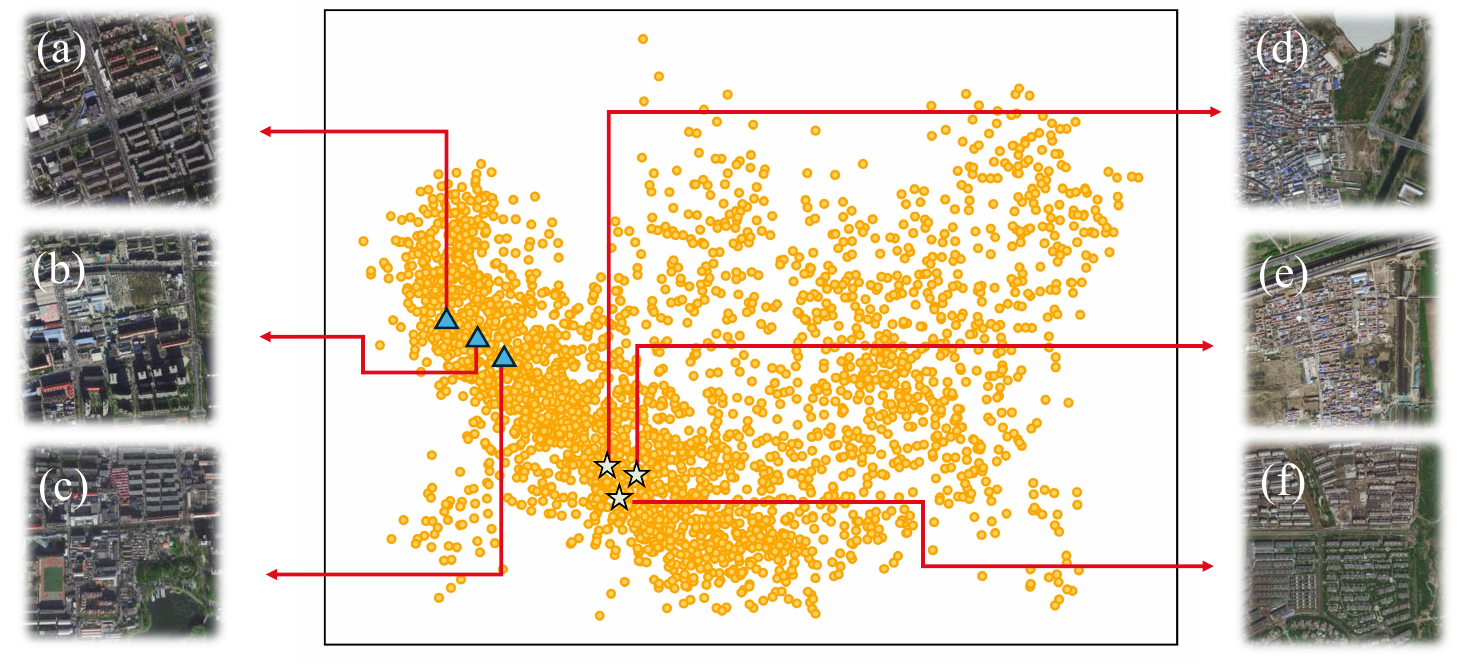}
\vspace{-0.5em}
    \caption{Representation space visualization.}
    \label{fig:cluster1}
    \vspace{-0.5em}
\end{figure}

In this section, we map the region representations learned in UrbanVLP into a two-dimensional space using the PCA algorithm in Figure~\ref{fig:cluster1}. 
As we can see, the three images on the left and on the right belong to different clusters, each exhibiting intra-cluster similarities and inter-cluster differences. Therefore, UrbanVLP could effectively model regions into high-dimensional space, wherein satellite images with similar architectural layouts demonstrate spatial similarity. Our framework learns a reasonable, accurate, and semantically rich region representation for USI prediction.

\begin{figure}[!h]
    \centering
\includegraphics[width=\columnwidth]{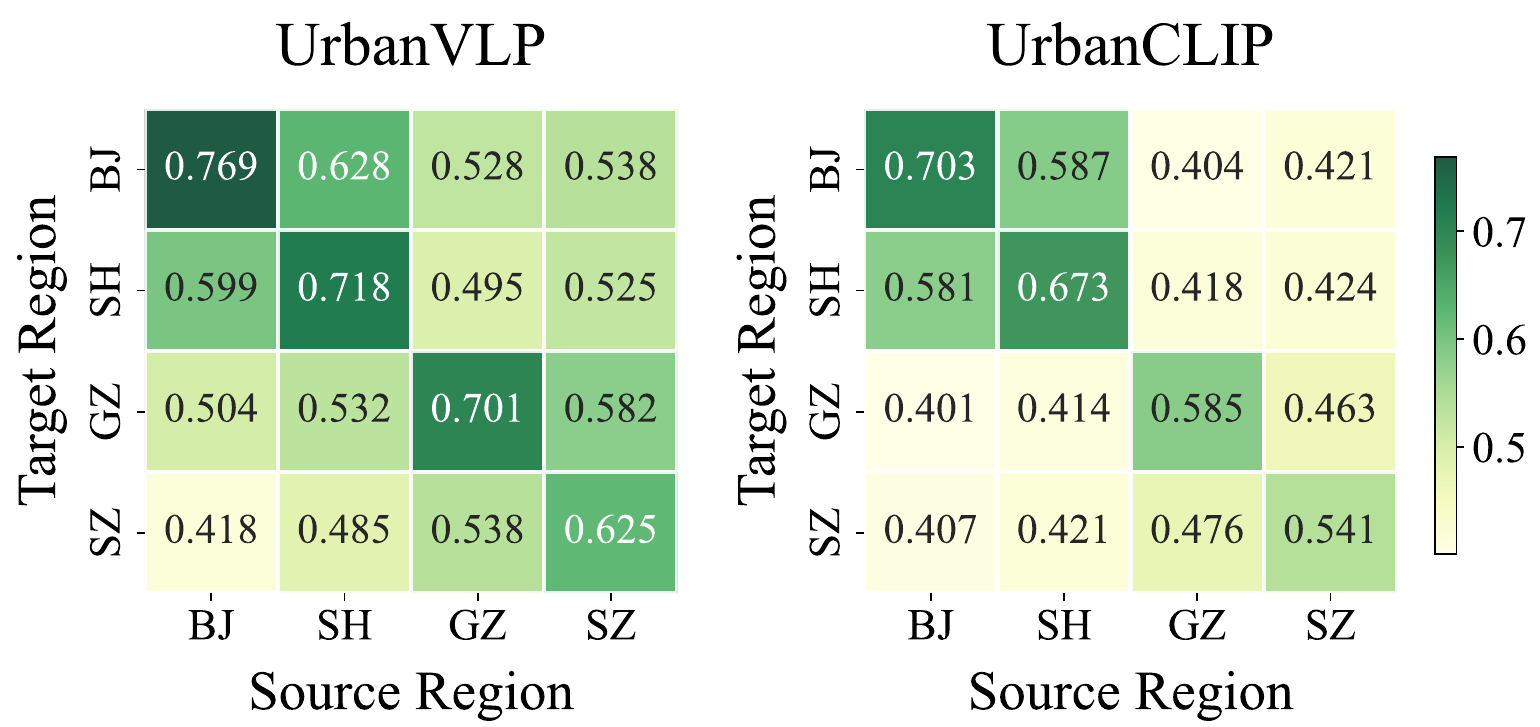}
    \caption{Transferability test on $R^2$ between UrbanVLP and UrbanCLIP, on the Carbon indicator across 4 cities.}
    \label{fig:transfer}
    \vspace{-1em}
\end{figure}
\subsection{Transferability Study}

In this study, we explore the practical application of USI prediction within a transfer learning context~\cite{park2022learning,liu2023knowledge}. 
Specifically, We use visual encoders trained on source city data and fine-tune them on target city data. 
We experiment with different pairs of source and target cities and present the carbon emission prediction in Figure~\ref{fig:transfer}.
As we can see, on 16 source-target city pairs, UrbanVLP achieves an average $R^2$ of 0.574, while that of UrbanCLIP is 0.495. 
Additionally, the $R^2$ values show greater similarity between Beijing and Shanghai, as well as between Shenzhen and Guangzhou. This observation aligns with the geographical proximity of Beijing to Shanghai and Guangzhou to Shenzhen.
These findings confirm the robust transferability of our UrbanVLP model in urban areas, despite the previously mentioned differences in geological and demographic characteristics among the selected cities.

\vspace{-0.5em}
\subsection{Analysis about Hyperparameters}

\begin{figure}[h!]
\centering
\includegraphics[width=\columnwidth]{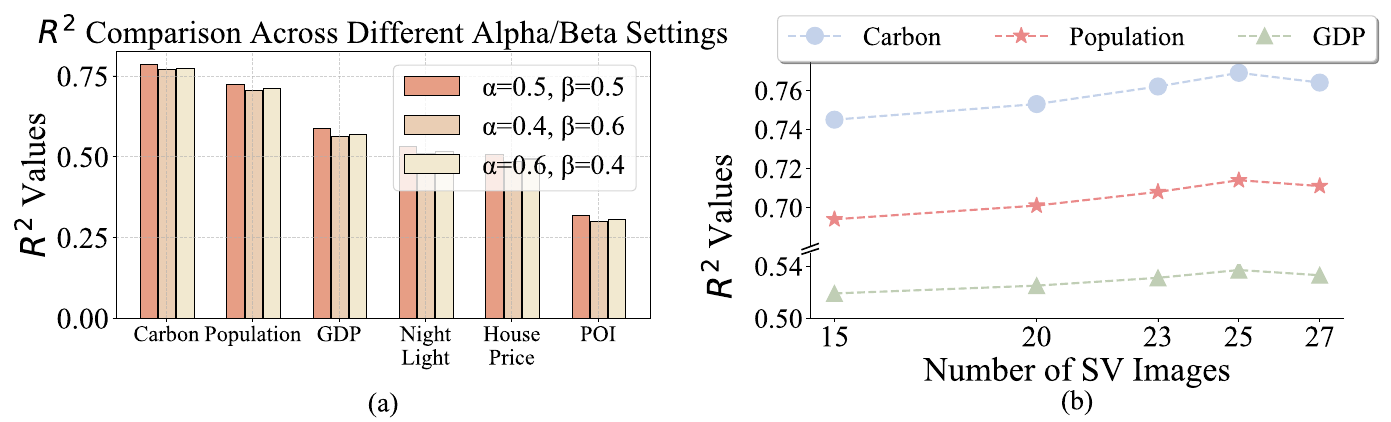}
    \vspace{-2em}
    \caption{(a) Hyperparameter for loss combination. (b) Experiments for the maximum number of street-view images. }
    \label{fig:combined_2plots}
    \vspace{-1em}
\end{figure}

In Figure \ref{fig:combined_2plots} (a), we explore different combinations of hyperparameter values for loss combination. It can be seen that the settings we choose in the paper ($\alpha$=0.5, $\beta$=0.5) yield the best experimental results.
We also study the maximum value of street-view images within one region during training, which is set as 25 in our paper. As is shown in Figure~\ref{fig:combined_2plots} (b), the $R^2$ value steadily increases with the number of street-view images, peaking at 25. Therefore, our setting is the optimal.

\subsection{Practicality} 
To showcase the practical applications, we further develop a web-based system (Figure~19 in Appendix), with more details and discussion in Appendix~K.
This system renders a visual representation of urban areas through satellite imagery, complemented by street-level photographs for an interactive exploration experience. 
\textit{It enhances social planning and sustainable development by providing a data-driven platform that deepens understanding of socioeconomic indicators}.

\vspace{-0.7em}
\section{Conclusion and Future Work}

USI prediction plays a significant role for understanding societal patterns and dynamics.
\model, for the first time, explores the differences between street-view and satellite images from a semantic granularity perspective, as well as their roles in modeling urban region representation. 
A text generation and calibration mechanism is also proposed to ensure high-quality description generation.
It has achieved state-of-the-art results on the constructed $\mathtt{\dataset}$ dataset. 
Future research could incorporate additional modalities such as POIs, road networks, and building footprints to enhance information richness and introduce a broader perspective.

\section{Acknowledgments}
This work is mainly supported by the National Natural Science Foundation of China (No.
62402414). This work is also supported by the Guangzhou-HKUST(GZ) Joint Funding Program (No.
2024A03J0620), Guangzhou Municipal Science and Technology Project (No. 2023A03J0011), the
Guangzhou Industrial Information and Intelligent Key Laboratory Project (No. 2024A03J0628), and
a grant from State Key Laboratory of Resources and Environmental Information System, and Guangdong Provincial Key Lab of Integrated Communication, Sensing and Computation for Ubiquitous
Internet of Things (No. 2023B1212010007).

\bibliography{aaai25}

\clearpage
\input{appendix/0_appendix_list}

\end{document}

%% file: appendix/0_appendix_list.tex
\input{appendix/formulation}
\input{appendix/related_work}
\input{appendix/datasets}

\input{appendix/motivation}
\input{appendix/baselines}
\input{appendix/implementation_detail}
\input{appendix/formulas_for_modaility_representation}
\input{appendix/segment_ratio}

\input{appendix/lmm}
\input{appendix/b_experimental}
\input{appendix/practicality}
\input{appendix/limitation_and_social_impact}

%% file: appendix/formulation.tex
\section{A ~~~Formulation}

\label{appendix:formulation}
\noindent \textbf{Definition 1 (Urban Region)} 
We follow previous research \cite{xi2022beyond,yan2023urban} to \emph{evenly} divide the target area (\eg~a city) into $L$ urban regions.

\vspace{0.2em}
\noindent \textbf{Definition 2 (Urban Imagery)} Diverse visual images, including satellite and street-view images, facilitate an intuitive analysis of urban regions. Specifically, \textit{satellite images} provide a coarse-grained overview of the structural composition of a region. Each input satellite image \wrt~the urban area $g$ can be represented as $I^{st}_g\in\mathbb{R}^{H\times W \times 3}$, where $H$ and $W$ denote the length and width. \textit{Street-view images} are captured by vehicles from map providers during urban traversals, offering detailed visual information on the fine-grained aspects within a region. By querying the latitude and longitude range of satellite images, multiple street-view images are associated together \wrt~the urban region $g$ based on their locations $\mathcal{L}_g$. Each individual street-view image is represented as $I^{{sv}}_g\in\mathbb{R}^{H\times W \times 3}$.

\vspace{0.2em}
\noindent \textbf{Definition 3 (Text Description)} Text descriptions of urban area $g$ (including satellite text $T^{st}_g$ and street-view text $T^{sv}_g$) provide precise tools for urban area analysis. Such texts can be manually generated or produced using image annotation. Particularly, leveraging advanced LLMs enables detailed and insightful analytical descriptions of given area images. 

\vspace{0.2em}
\noindent \textbf{Definition 4 (Urban Socioeconomic Indicators)} Urban socioeconomic indicators assess various aspects of a region, including population, economy, society, and public services, and so on. 
A set of $L$ urban regions with $K$ indicators is represented as $\mathbf{Y} \in \mathbb{R}^{L \times K}$. In this paper, we comprehensively use \emph{population}, \emph{GDP}, \emph{carbon emissions}, \emph{night lights}, \emph{house prices} , and \emph{points of interest}, \ie~\emph{POI}, as ground-truth urban indicators.

%% file: appendix/related_work.tex
\section{B ~~~Detailed Related Work}

\noindent \textbf{Urban Socioeconomic Indicator Prediction.} 
In literature, many studies have focused on learning task-specific region representations from various urban data,
especially urban imagery due to its consistent updates and easy accessibility \cite{10.1145/3511808.3557153,liu2023knowledge,xi2022beyond}.
For example, Urban2Vec \cite{wang2020urban2vec} integrated \emph{street-view imagery} and POI data to learn neighborhood embeddings. Some contrastive learning approaches like PG-SimCLR \cite{xi2022beyond} and UrbanCLIP \cite{yan2023urban} have shown success in representing urban regions through \emph{satellite images}.
However, the aforementioned works exclusively considered a single type of urban imagery, \emph{overlooking the potential synergy between satellite imagery and street-view imagery, which can complement each other.}
Recently, some efforts \cite{rs12020329,HUANG2023102043,10.1145/3511808.3557153,CAO2023103323,10.1145/3342240,cao2018integrating,HUANG2023102043} have begun to explore the simple combination of satellite imagery and street-view imagery. 
Vision-LSTM \cite{HUANG2023102043} integrates diverse street-level images into the satellite image domain using an LSTM-based network for precise urban village identification. \citet{10.1145/3511808.3557153} propose utilizing street segments for region modeling and predicting socioeconomic indicators by merging street-view and satellite images through a multi-stage pipeline.
Nevertheless, prior studies mostly overlook distinct features in satellite and street-view images for varied granularities. To this end, we propose a unified multi-granularity cross-modal alignment framework for comprehensive urban representation in this study.

\vspace{0.5em}
\noindent \textbf{Vision-Language Pretraining (VLP).} 
VLP aims to jointly encode vision and language in a fusion model.
Early works on VLP can be broadly categorized into single-stream \cite{chen2020uniter,kim2021vilt,li2020unicoder} and two-stream \cite{Tan2019LXMERTLC,luViLBERT} methodologies, with the former using a unified architecture for modality integration and the latter employing separate encoders before merging modalities. 
Milestone work CLIP \cite{radford2021learning} and its variants \cite{li2021supervision,yao2022filip} highlight the efficacy of contrastive learning in cross-modal downstream tasks, such as zero-shot learning and cross-modal retrieval.
Recent works \cite{tsimpoukelli2021multimodal,alayrac2022flamingo} shift towards leveraging LLMs knowledge for vision-language representation learning. \emph{In urban region profiling, the potential of VLP paradigm remains untapped, with limited exploration of the benefits of textual information.} Our groundbreaking work introduces multiple types of text descriptions as supplements, aiming to learn more comprehensive and interpretable urban region representations.

\vspace{0.5em}
\noindent \textbf{Large Multimodal Model (LMM).} 
As LLMs undergo rapid evolution, a faction within the research community is increasingly focused on integrating visual knowledge into these models, such as CLIP \cite{radford2021learning} and ALIGN \cite{jia2021scaling}. Early studies \cite{li2022blip,li2023blip} enhanced CLIP with refined data strategies for more diverse datasets, which have proven effective for fundamental visual tasks \cite{li2022grounded,liu2023grounding,zhang2022glipv2} but have shown limitations for complex tasks such as visual question answering. Recently, new models like MiniGPT-4 \cite{chen2023minigpt}, LLaVA \cite{liu2023visual}, and InstructBLIP \cite{instructblip} have improved the understanding of complex problems by expanding parameters and training data, called large multimodal models (LMMs). 
In this paper, we comprehensively evaluate the text generation quality of representative LMMs model and propose an automated text description generation and calibration technique to ensure effective image description capabilities with LMMs.

%% file: appendix/datasets.tex
\section{C ~~~More Details about Datasets \& Tasks}
\label{appendix:dataset}

\subsection{\dataset Dataset}
\begin{figure}[h]
    \centering
\includegraphics[width=0.8\columnwidth]{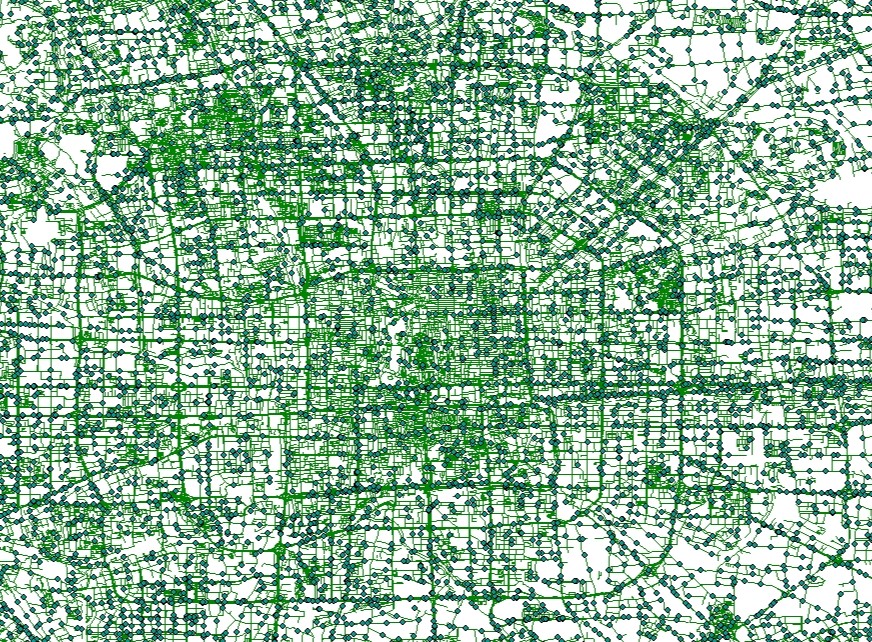} 
    \caption{Road Network Sampling for Street-View Images on Beijing.}
    \label{fig:road_network_sampling}
\end{figure}

We construct the $\mathtt{\dataset}$ Dataset using street-view geo-tagged images collected from Baidu Map API following previous works~\cite{liu2023knowledge} and generated text descriptions by LMM. We collect and generate satellite data following the practice in ~\cite{yan2023urban}.
\begin{itemize}
\item \textbf{Street-View Panoramic Image Collection}.
The Street-View image sampling process utilized the Baidu Map API\footnote{\url{https://lbsyun.baidu.com/}}, after the extraction of road network data within our predefined sampling area. Using the ArcGIS\footnote{\url{https://www.esri.com/en-us/arcgis/products/develop-with-arcgis/overview}} tool, we first exported the road network data. We then set a sampling interval of 500 meters, allowing for the collection of street-view images at every half-kilometer point along the road network. This approach ensured thorough coverage of the corresponding area, as demonstrated in Figure~\ref{fig:road_network_sampling}.
Based on the sampling points, we collected panoramic street-view images at a resolution of 2048x664, which significantly exceeds the 256x256 resolution of satellite imagery~\cite{yan2023urban}. This higher resolution ensures greater clarity and provides ample detail for further analysis.

As is shown in Table \ref{tab:dataset_statistics}, a total of 47,582 street-view images were collected, whose data volume is 2.7 times larger than that of the satellite data, capturing the intricacies of the urban landscape and providing a street-level view that enhances the overall urban analysis.


\item \textbf{Corresponding Text Description Generation}.
The textual data corresponding to each street-view image is generated from ShareGPT4V \cite{chen2023sharegpt4v}, which is recognized for its superior text generation capabilities compared to other contemporary open-source LMM models \cite{liu2023improved,instructblip,ye2023mplug,gao2023llama,chen2023minigpt,chen2023sharegpt4v}.
The quality of the generated data is guaranteed by our proposed \texttt{PerceptionScore}. We filter out the texts with \texttt{PerceptionScore} lower than 0.6.
\end{itemize}

\begin{table}[h]
\centering
\small
\caption{Urban imagery dataset statistics.}
\label{tab:dataset_statistics}
\scalebox{0.7}{
\begin{tabular}{l|c|c|c|c} 
\hline
\multirow{2}{*}{\textbf{Dataset}} & \multicolumn{2}{c|}{\textbf{Coverage}} & \multirow{2}{*}{\begin{tabular}[c]{@{}c@{}}\textbf{\#Street-view}\\\textbf{Image}\end{tabular}} & \multirow{2}{*}{\begin{tabular}[c]{@{}c@{}}\textbf{\#Satellite}\\\textbf{Image}\end{tabular}}  \\ 
\cline{2-3}
& \begin{tabular}[c]{@{}c@{}}\textbf{Bottom-left}\end{tabular} & \begin{tabular}[c]{@{}c@{}}\textbf{Top-right}\end{tabular} && \\ 
\hline
Beijing & 39.75°N, 116.03°E & 40.15°N, 116.79°E & 14,164 & 4,592 \\
Shanghai & 30.98°N, 121.10°E & 31.51°N, 121.80°E & 15,616& 5,244 \\
Guangzhou & 22.94°N, 113.10°E & 23.40°N, 113.68°E & 7,635& 3,402 \\
Shenzhen & 22.45°N, 113.75°E & 22.84°N, 114.62°E & 10,167&4,324 \\
\hline
\end{tabular}}
\vspace{-2mm}
\end{table}

\subsection{Downstream Dataset \& Tasks}

In our research, the Downstream Dataset plays a crucial role in gauging the practical implications of our multimodal framework. 
We collect six representative urban socioeconomic indicators:  \emph{Carbon Emission, Population, GDP, Night Light, House Price}, and \emph{POI}. Each indicator illuminates a significant dimension of urban vitality and sustainability.
\input{appendix/downstream_indicators}


\subsection{Downstream Dataset Analysis}

\begin{figure}[h]
    \centering
    \includegraphics[width=1.0\columnwidth]{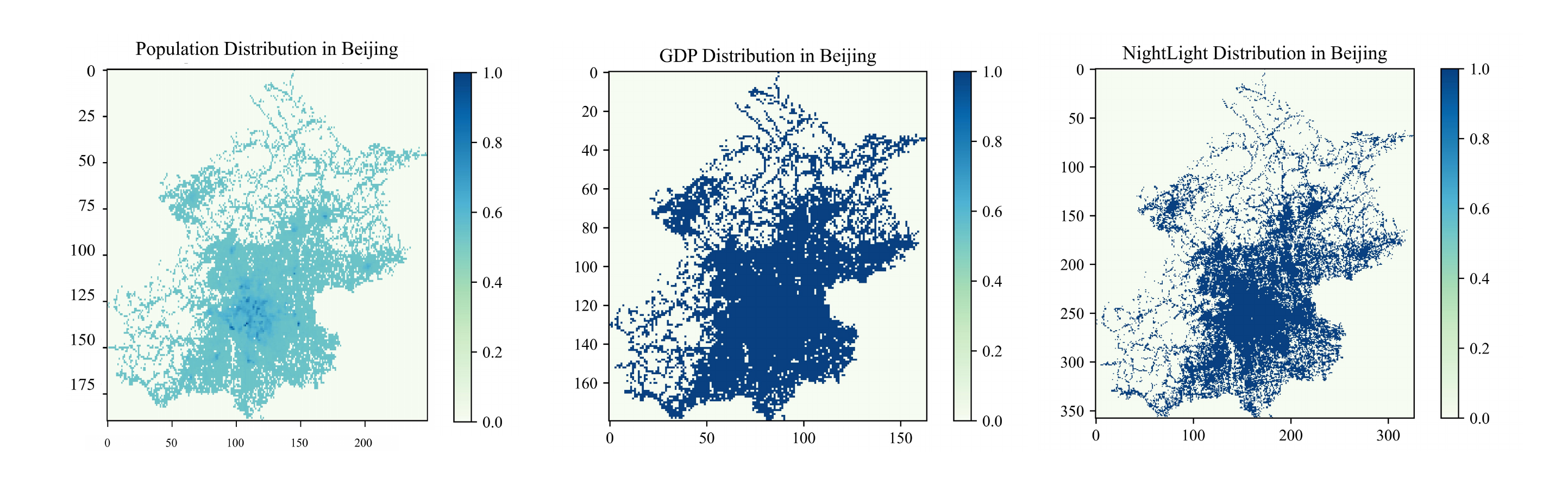}
    \caption{Spatial distribution of Popultion, GDP and NightLight in Beijing. }
\label{fig:downstream_distribution}
\vspace{-1em}
\end{figure}

To provide readers with an intuitive understanding, we visualize the distribution of three downstream socioeconomic indicators for Beijing in Figure~\ref{fig:downstream_distribution}.
The distribution of different indicators exhibits similar trends, indicating that Beijing has a clear ``center-periphery'' structure, with the core area being densely populated and economically developed, while the peripheral areas have relatively sparse economic and demographic data.


\vspace{-1em}

\begin{figure}[h]
    \centering
\includegraphics[width=0.8\columnwidth]{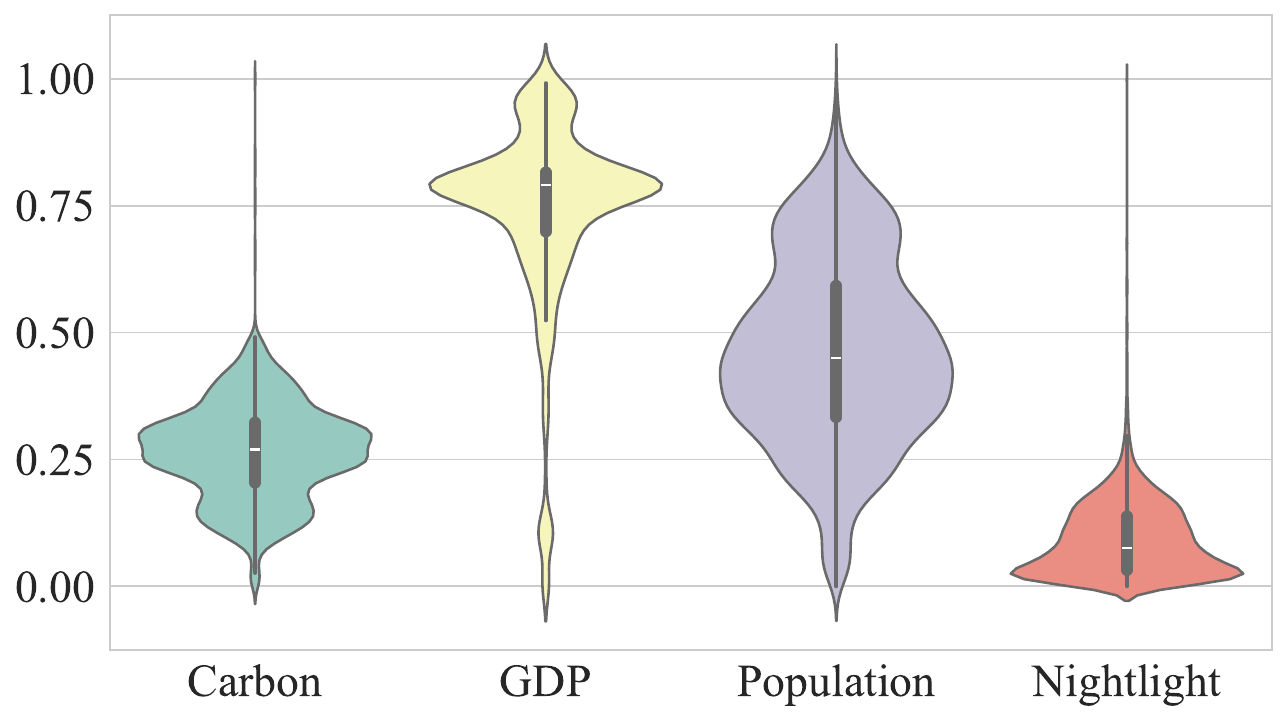}
 \caption{Data distribution of Carbon Emission, Popultion, GDP and Night Light in Beijing.}
        \label{fig:boxplot}
\end{figure}

Beyond geographical spatial differences, the concentration trends and variability of the data are critical perspectives that merit attention.
In Figure~\ref{fig:boxplot} we demonstrate these patterns by illustrating the normalized values of different socioeconomic indicators in Beijing.
Outliers with extreme maximum and minimum values will be filtered out during our pre-processing.
As we can see, the values of Carbon Emission and Night Light are relatively concentrated, showing minimal variability. The distribution of GDP values is more widespread, while Population distribution exhibits a bimodal characteristic, reflecting the inherent diversity.

%% file: appendix/downstream_indicators.tex
In this section, we provide the detailed introduction and data source of our downstream benchmark.

\begin{itemize}
    \item \textbf{Carbon Emissions:} 
    Extracted from the Open-source Data Inventory for Anthropogenic CO\textsubscript{2} (ODIAC), this dataset offers granular carbon emission figures in 2022 that are in alignment with the spatial resolution of our satellite imagery, which is one square kilometer per image. The emissions are quantified in tons on a monthly basis.
    
    \item \textbf{Population:} 
    The quantum of population serves as an indicator of a region's sociodemographic characteristics.
    Sourced from WorldPop (\url{https://hub.worldpop.org}), we gather population distribution data in 2020, to provide an in-depth portrayal of demographic patterns.
    The unit is \#citizens.        
    
    \item \textbf{GDP:} 
    Our dataset encapsulates China's economic development by incorporating Gross Domestic Product (GDP) figures, thereby infusing an economic dimension into our analysis of urban regions. The unit is million Chinese Yuan.

    \item \textbf{Night Light:}
    Nighttime light can, to some extent, characterize the intensity of human activities, which is of great significance for urban development studies.
    We take the nightlight data from \cite{zhong2022long} in 2020.
    \item \textbf{House Price:}
    The price of housing is subject to the influence of numerous socioeconomic factors. By observing fluctuations in housing prices, it is possible to discern, to a certain extent, the state of economic development.
    We collect data from Anjuke (\url{https://www.anjuke.com}) in March 2023 with yuan/$m^2$ unit. 
    Since individual house prices represent locational attributes, we extrapolate a regional-level characteristic by averaging housing prices across an urban region.
    
    \item \textbf{POI:}
    The spatial distribution of various types of POIs (Points of Interest) within urban areas is a potent social indicator, as it reflects the underlying social and economic architecture, as well as the population's activities and urban lifestyles. Consequently, to glean insights into these urban dynamics, we collect data of POIs from web sources, encompassing both their locations and categories.
    The collected data has 14 categories: \emph{Dining and Cuisine; Leisure and Entertainment; Sports and Fitness; Business and Residential; Healthcare; Financial Institutions; Tourist Attractions; Lifestyle Services; Shopping and Consumption; Automobile Related; Hotel Accommodation; Transport Facilities; Science, Education and Culture; Companies and Enterprises}.
    To capture the unique attributes of each area, we aggregate the number of POIs by category within each region.
    
\end{itemize}

In accordance with established conventions~\cite{liu2023knowledge,yan2023urban,xi2022beyond}, we uniformly and randomly partition the downstream dataset into train:val:test=7:1:2.
The data preprocessing setting adheres to \cite{yan2023urban} with refinement.
For some indicators data with relatively small differences in values, we have taken the logarithm of these values to facilitate training.

%% file: appendix/motivation.tex
\section{D ~~~Motivation Demonstration for Multi-granularity Modeling}
\begin{figure}[h]
    \centering
    \includegraphics[width=\linewidth]{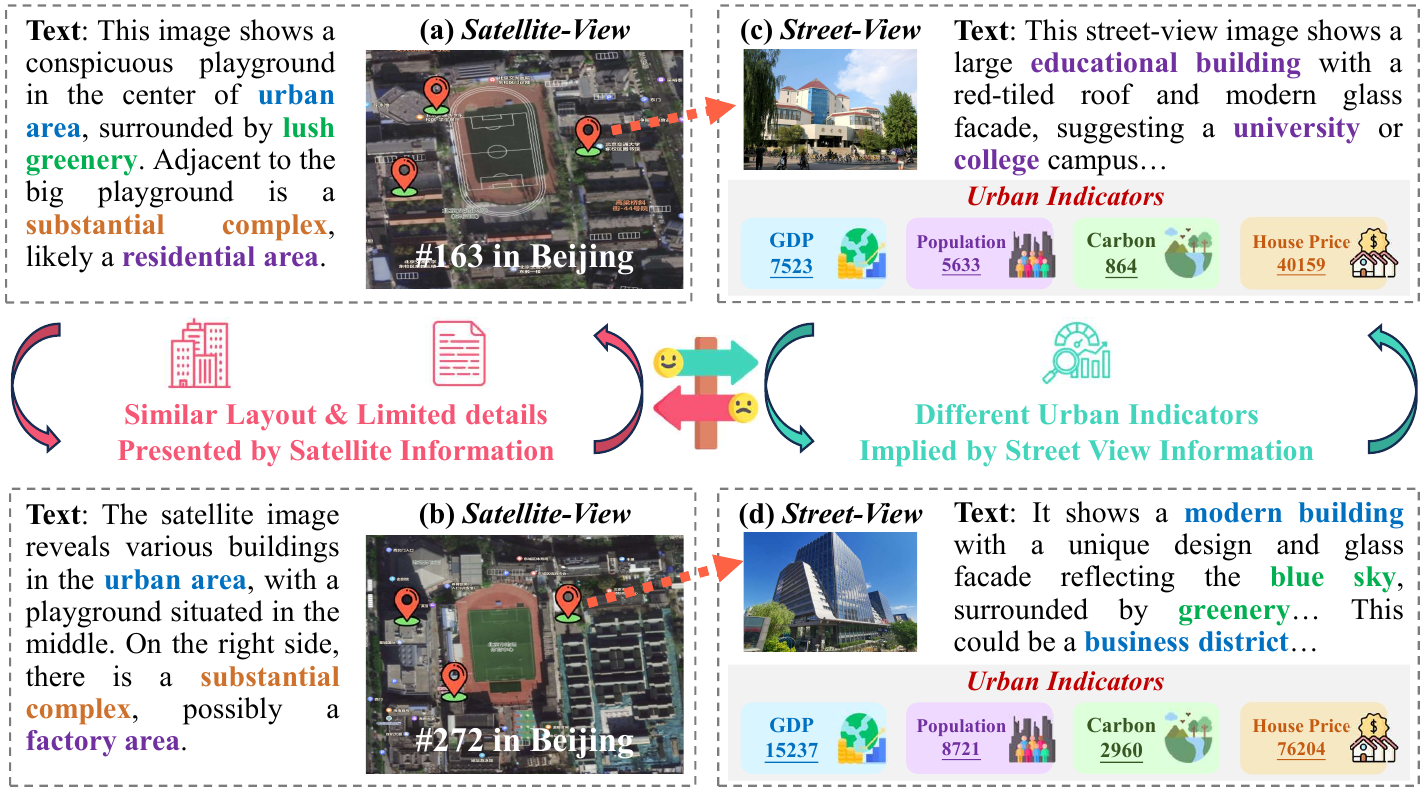}
    \vspace{-1.5em}
    \caption{Single-granularity vs. Multi-granularity modeling.}\vspace{-1em}
    \label{fig:mot}
\end{figure}

Here, we provide a more detailed example to illustrate our motivation for urban multi-granularity modeling.
Urban environments exhibit a spatial hierarchy in reality,
from a macro region level to a micro location level (e.g., architectural
details, street furniture, storefront). As depicted in Figure \ref{fig:mot}, while
the satellite images (a) and (b) possess remarkably similar layouts,
their urban indicators considerably differ. Upon zooming into micro
levels using the corresponding street-view images (c) and (d), a more nuanced and fine-grained understanding becomes apparent. 
We can observe that their differences arise from the varying surrounding environments (i.e., educational buildings vs. business districts).
Exclusively capturing macro-level patterns through satellite imagery
may introduce substantial bias, falling short in furnishing more
comprehensive information across diverse urban levels.

%% file: appendix/baselines.tex
\section{E ~~~In-depth Overview of Baseline Methods}
\label{appendix:baseline}
The single granularity imagery-based USI prediction methods include:

\begin{itemize}
  \item \textbf{ViT \cite{dosovitskiy2021an}}: ViT is a successful paradigm of applying transformers to computer vision, where images are segmented into patches for fixed-length modeling. When sufficient data is available for pre-training, ViT's performance can surpass that of CNNs~\cite{he2016deep}. In our experiment, we leverage ViT-B as the baseline for comparison.
  
  \item \textbf{PG-SimCLR \cite{xi2022beyond}}: 
  A contrastive learning framework which introduces geographic information (i.e., POI) into urban region representation learning. 
  In our study, we do not employ PG-SimCLR to predict POI to ensure a fair comparison.
  \item \textbf{UrbanCLIP \cite{yan2023urban}}: A model that utilizes an image-text contrastive learning-based approach for learning robust representations from urban imagery, intending to capture the complexity and diversity of urban areas.
  \item \textbf{UrbanCLIP-SV \cite{yan2023urban}}: Original UrbanCLIP is pretrained on satellite image-text pairs. Here we also verify its performance on street-view image-text pairs of our proposed $\mathtt{\dataset}$ dataset, 
   utilizing the same aggregation method to acquire regional features.
\end{itemize}

We also apply multiple baselines for multi-granularity USI prediction:

\begin{itemize}
    \item \textbf{Vision-LSTM \cite{HUANG2023102043}}:  A vision long short-term memory neural network which could process variable-length street-view image within each region. 
    The integration of satellite and street-view imagery, along with trajectories, facilitates a more effective capture of comprehensive urban representations, yielding reliable recognition results. 
    In our experiments, we include only satellite and street-view data, without incorporating trajectory data.
    \item \textbf{StructuralUrban~\cite{10.1145/3511808.3557153}}: The graph-based framework for urban region profiling leveraging street segments as containers to adaptively fuse the features of multi-level urban images. We term this method StructuralUrban for simplicity.
\end{itemize}


%% file: appendix/implementation_detail.tex
\section{F ~~~Implementation Details}
\label{appendix:implement}
In our experiment, Adam optimizer~\cite{adam2015} is chosen to minimize the training loss during the parameter learning. A grid search on hyperparameters is conducted, where search ranges for learning rate and batch size are set as 
{2e\mbox{-}6, 2e\mbox{-}5, 2e\mbox{-}4, 2e\mbox{-}3, 2e\mbox{-}2} and {4, 8, 16, 32, 64}, respectively. The data augmentation strategy follows the setting of \cite{radford2021learning}.
The weight hyperparameter $\alpha$ and $\beta$ of $\mathcal{L}_{Total}$ are set to 0.5, 0.5 respectively.
We run all the models on NVIDIA A800 GPUs with PyTorch 1.13.1 on Ubuntu 22.04. The number of epochs for the pretraining and fine-tuning stages is set to 10 and 100, respectively, with an early stopping strategy. 

We implement the aggregation method $\operatorname{Aggr}(\cdot)$ through MLP layers.
The feature fusion method $\textit{f}~$ directly utilizes additive aggregation (the experiments regarding the choice of fusion method are provided in Appendix~J).
To ensure a fixed length for network processing, we set the maximum street-view sequence length within one region as 25, and sequences shorter than that are achieved by padding with zeros.
For the semantic segmentation model to decouple visual elements with street-view images, following the successful precedent set by ~\cite{fan2023urban}, the architecture for the semantic model is \textit{ResNet18dilated + PPM\_deepsup}.
Initially, the dataset comprises 150 segmentation categories. However, in line with our requirements, we narrow down the selection to 38 categories, which we further group into 13 street-view features following \cite{fan2023urban}: \emph{Person; Bike; Heavy Vehicle; Light Vehicle; Facade; Window \& Opening; Road; Sidewalk; Street Furniture; Greenery - Tree; Greenery - Grass \& Shrubs; Sky; Nature}.

%% file: appendix/formulas_for_modaility_representation.tex
\section{G ~~~More Details for Modality Representation}
For utilizing ViT as the visual encoder to process urban visual data, we reshape the input image $I \in \mathbb{R}^{H\times W \times 3} $ into a sequence of flattened 2D patches ${I}_P \in \mathbb{R}^{N\times ({P^{2}} \cdot {C})}$, where C represents the number of channels, $(P,P)$ specifies the resolution of each image patch, and $N = HW/P^2$ denotes the number of patches, corresponding to the input sequence length for ViT. Notebly, the ViT utilizes a consistent latent vector size (denoted as $d$) across all its layers. Thus, we can obtain the outputs as the patch embeddings through a trainable linear projection. We also prepend a learnable embedding to the patch embedding sequence to serve as the overall representation (i.e., $\mathbf{z}_{0}^{0}={I}_{cls}$). Besides, we add learnable position embedding $\mathbf{E}_{pos}$ to retain positional information. The process is as follows:
\begin{equation}\label{eq:vit1}
    \mathbf{z}_0=[{I}_{cls}; {I}_p^1\mathbf{W}; {I}_p^2\mathbf{W}; \cdots; {I}_p^N\mathbf{W}]+\mathbf{E}_{pos},
\end{equation}
where $\mathbf{W}\in\mathbb{R}^{(P^2\cdot C)\times d}$ 

Consistent with the standard Transformer encoder, ViT includes alternating layers of multi-head self-attention $\operatorname{MSA(\cdot)}$ operation \cite{vaswani2017attention}. Furthermore, layer normalization (LN) is adopted before every block, whereas residual connections are applied after every block:
\begin{equation}\label{eq:vit2}
    \mathbf{z}_l = \mathrm{MSA}(\mathrm{LN}(\mathbf{z}_{l-1}))+\mathbf{z}_{l-1}, \quad l=1\ldots L.
\end{equation}

The text encoder follows a similar $\operatorname{MSA}$ mechanism of the visual encoder. The input text sequence is bracketed with $[SOS]$ and $[EOS]$ tokens, and the activation of the highest layer of Transformer at $[EOS]$ token is considered the global representation $\mathbf{z}_T$ of text.

%% file: appendix/segment_ratio.tex
\section{H ~~~Segmentation Ratio and Prompt Examples}
\label{appendix:segmentation_ratio}
\begin{figure}[h]
    \centering
    \includegraphics[width=0.9\columnwidth]
    {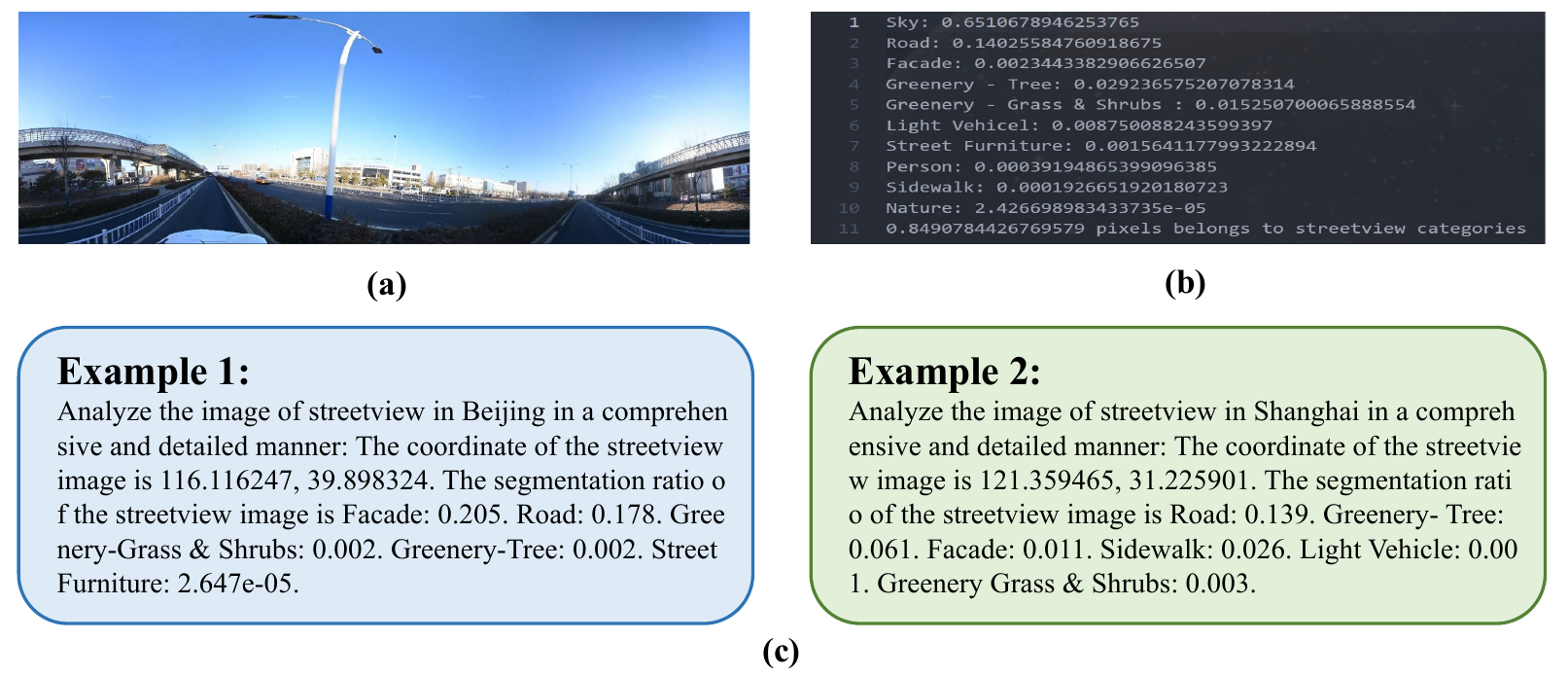}
     \vspace{-2mm}
    \caption{Prompt Examples. 
    }
    \vspace{-4mm}
    \label{fig:prompt_example}
\end{figure}

As illustrated in Figure \ref{fig:prompt_example} (a) and (b), we demonstrate examples of the segmentation ratio of street-view images in our $\mathtt{\dataset}$ dataset. The segmentation ratio is decoupled by pretrained segmentation models and is used in both prompt construction and \texttt{PerceptionScore} computation.
We group the segmentation categories into 13 categories following~\cite{fan2023urban}: \emph{Person; Bike; Heavy Vehicle; Light Vehicle; Facade; Window \& Opening; Road; Sidewalk; Street Furniture; Greenery - Tree; Greenery - Grass \& Shrubs; Sky; Nature}.

The prompt template for generating street-view descriptions is formulated as follows: \emph{Analyze the street-view panoramic image in [City] in a comprehensive and detailed manner. The coordinate of the street-view image is [Longitude, Latitude]. The segmentation ratio of the street-view image is [Segment Information]}.
Figure \ref{fig:prompt_example} (c) provides illustrations of some examples of our prompts for description generation.


%% file: appendix/lmm.tex
\section{I ~~~The Choice of Image-to-Text LMM}
\label{appendix:lmms}
In our work, we leverage ShareGPT4V \cite{chen2023sharegpt4v} for text description generation. Here we conduct experiments on current open-sourced LMM models, and compute the \texttt{PerceptionScore} of them. The results are presented in Table \ref{table:lmms}.
As we can see, ShareGPT4V achieves the best performance in terms of \texttt{PerceptionScore}, followed by LLaMA-Adapter V2, which is used as the Image-to-Text models in~\cite{yan2023urban}.
Additionally, we note that there is study~\cite{zhong2024urbancross} argue that InstructBLIP~\cite{instructblip} exhibits state-of-the-art (SOTA) performance in image-to-text generation. However, in our experiment, we found that InstructBLIP is unsuitable for handling geospatial coordinates as prompt content.
Incorporating such coordinates significantly degrades the quality of the generated text, leading to poor content output. Due to these limitations, we have excluded it from our comparative analysis.

\begin{table}[h]
\centering
\caption{\texttt{PerceptionScore} of representative LMMs. GD is short for Generated-Description.}
\setlength{\tabcolsep}{2pt} 
\renewcommand{\arraystretch}{1.2} 
\begin{tabular}{@{}lc@{}}
\toprule
Models & Average GD-Scores \\ 
\midrule
LLaVA-v1.5 \cite{liu2023improved}  & 0.674 \\
mPLUG-Owl \cite{ye2023mplug} & 0.662 \\
LLaMA-Adapter V2 \cite{gao2023llama} & 0.698 \\
MiniGPTv2 \cite{chen2023minigpt} & 0.693 \\
\hline
ShareGPT4V \cite{chen2023sharegpt4v} & 0.714 \\
\bottomrule
\end{tabular}
\label{table:lmms}
\vspace{-1em}
\end{table}

%% file: appendix/b_experimental.tex
\section{J ~~~More Experimental Analysis}
\label{appendix:exp}

\input{appendix/text_calibration_new}

\subsection{Analysis about Time Complexity}
Researches in urban areas can provide valuable references and assistance to policymakers, builders, and citizens. Therefore, the practicality of the model should be taken into consideration.

\begin{figure}[h!]
\centering
\includegraphics[width=\columnwidth]{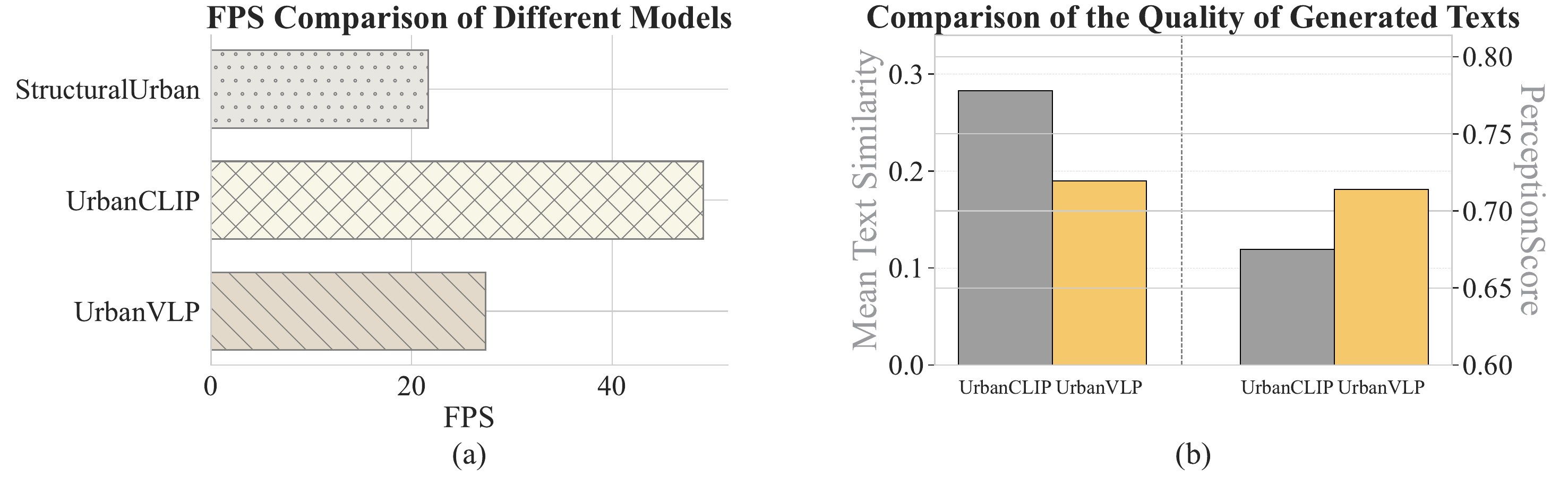}
    \vspace{-1.3em}
    \caption{(a) FPS comparison of different models. (b) Comparison of the quality of generated texts of UrbanVLP and UrbanCLIP.}
    \label{fig:combined_plotscd}
    \vspace{-2em}
\end{figure}

We conduct time complexity analysis using FPS (Frames Per Second), a common indicator for inference speed in computer vision area. The result is reported in Figure \ref{fig:combined_plotscd} (a). 
In fact, compared to UrbanCLIP, UrbanVLP's introduction of street-view features for improved and finer-grained effects will inevitably sacrifice a portion of the time efficiency. However, compared to StructuralUrban's graph structure inference, our linear probing setting ensures the minimum time overhead, also with better performance.
For making model decisions during deployment, to pursue the trade-off between performance and speed, UrbanVLP is the optimal choice.
We also present a practical deployment example of UrbanVLP in Appendix~K.

\subsection{Discussion on the Enhancements in Text Generation Quality by UrbanVLP}
As we discussed in the Introduction,
the text generation process of UrbanCLIP raises several critical concerns: i) Hallucination and ii) Homogenization.
To assess the comparative improvement of UrbanVLP in addressing these issues, we calculate three metrics for the generated text in both UrbanCLIP and UrbanVLP: (1) the average text length, (2) the mean text similarity, and (3) the average \texttt{PerceptionScore}.

First of all, the average length of the text description in UrbanCLIP is quantified at 18.1 words. This relatively short length may lead to insufficient information being conveyed. In contrast, the average length of the text description in UrbanVLP is 144.1 words, nearly eight times longer, providing ample space to describe the content of the images in greater detail, thereby covering more nuanced information.
Furthermore, we calculate the mean text similarity of all the generated texts of UrbanCLIP and UrbanVLP. As depicted in Figure~\ref{fig:combined_plotscd} (b) on the left side, the mean text similarity of UrbanCLIP texts is 0.28, while that of UrbanVLP is 0.19. This indicates that our Automatic Text Generation and Calibration mechanism significantly mitigates the issue of homogenization, resulting in more diversified text descriptions for images with different content. 
Additionally, on the right side of Figure~\ref{fig:combined_plotscd} (b), we evaluate the quality of generated descriptions through our proposed \texttt{PerceptionScore}. As we can see, the average scores of UrbanCLIP texts and UrbanVLP texts are 0.675 and 0.714, respectively. 
This indicates that UrbanVLP effectively mitigates the issue of inconsistency (hallucination) between generated text and visual content.

\subsection{Ablation study for fusion method}
To verify the effectiveness of our addition fusion method of satellite image features, street-view image features and location features, we
experiment with different types fusion method, including (1) Concatenation: we directly concat the three kinds of features; (2) MLP: we feed the three features into a multi-layer perceptron (MLP) for fusion, resulting in a single output. As indicated in Table~\ref{tab:ablationfusion},
 the differences among several distinct fusion methods are relatively minor, which may be attributed to the overall complexity of the model, resulting in minimal impact from the various fusion techniques. Consequently, we opt for the most straightforward and high-performing approach, the addition method, for implementation.

\vspace{-0.5em}
\begin{table}[!htbp]
\centering
\caption{Ablation study on $\mathtt{\dataset}$-$\mathtt{Beijing}$  dataset. The best results are in bold.}
\vspace{-2mm}
\setlength{\tabcolsep}{0.8mm}
\scalebox{0.7}{
\begin{tabular}{l|ccc|ccc|ccc}
\hline
\multirow{2}{*}{\textbf{Model}}
& \multicolumn{3}{c|}{\textbf{Carbon}} & \multicolumn{3}{c|}{\textbf{Population}} & \multicolumn{3}{c}{\textbf{GDP}}  \\

& $\text{R}^2$ & $\text{RMSE}$ & $\text{MAE}$ & $\text{R}^2$ & $\text{RMSE}$ & $\text{MAE}$ & $\text{R}^2$ & $\text{RMSE}$ & $\text{MAE}$  \\
\hline

Concat & 0.759 &     0.487  & 0.389 &0.705    &0.535  & 0.429&  0.526   & 0.693   & 0.424 

\\
MLP &0.761&	0.483	&0.374&0.710&	0.527&	0.415 & 0.532	&0.688	&0.418
\\
Addition (Ours) & \textbf{0.769} & \textbf{0.477} & \textbf{0.369}     & \textbf{0.714} &
\textbf{0.523}&
\textbf{0.411}&
\textbf{0.537}&
\textbf{0.684}&
\textbf{0.416}
\\
\hline
\end{tabular}
}
\label{tab:ablationfusion}
\end{table}

\subsection{More Visualization of Region Representations}
\begin{figure}[h]
    \vspace{-1em}
    \centering
\includegraphics[width=0.95\columnwidth]{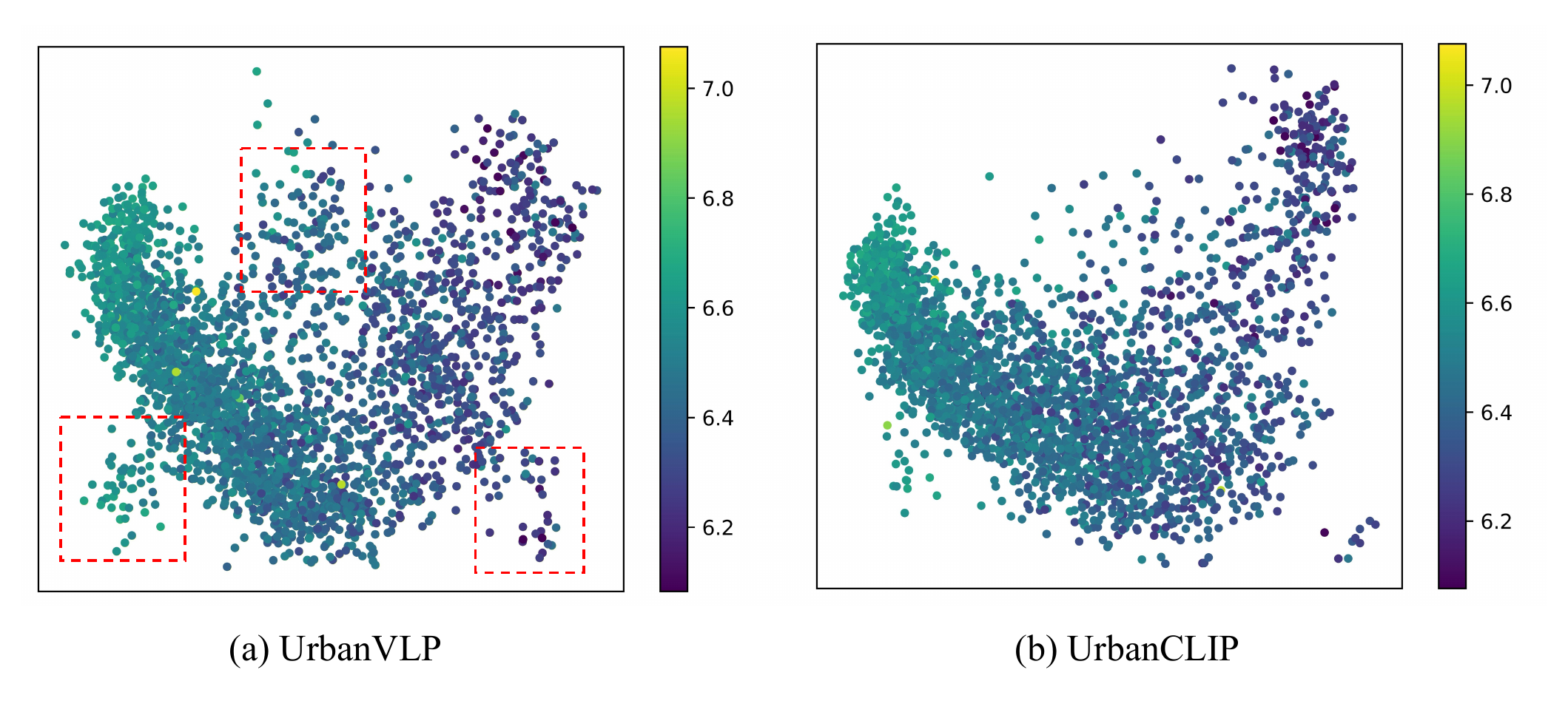}
    \vspace{-0.5em}
    \caption{PCA visualization of urban imagery.}
    \label{fig:cluster2}
\end{figure}

Figure~\ref{fig:cluster2} provides a visualization with a population indicator, compared to UrbanCLIP. We use the dot color to indicate the population (log) in the corresponding regions of Beijing.
As shown by the red dashed boxes, the regions modeled by UrbanVLP exhibit more significant clustering characteristics and greater distinction in high-dimensional space. 
Additionally, the lower parts of the two main band-shaped branches are also more distinctly separated.
Points that are spatially close also have similar population values.

%% file: appendix/text_calibration_new.tex
\subsection{More Analysis for Text Calibration}
\label{appendix:calibration}
\begin{figure}[h]
    \centering
\includegraphics[width=0.8\columnwidth]{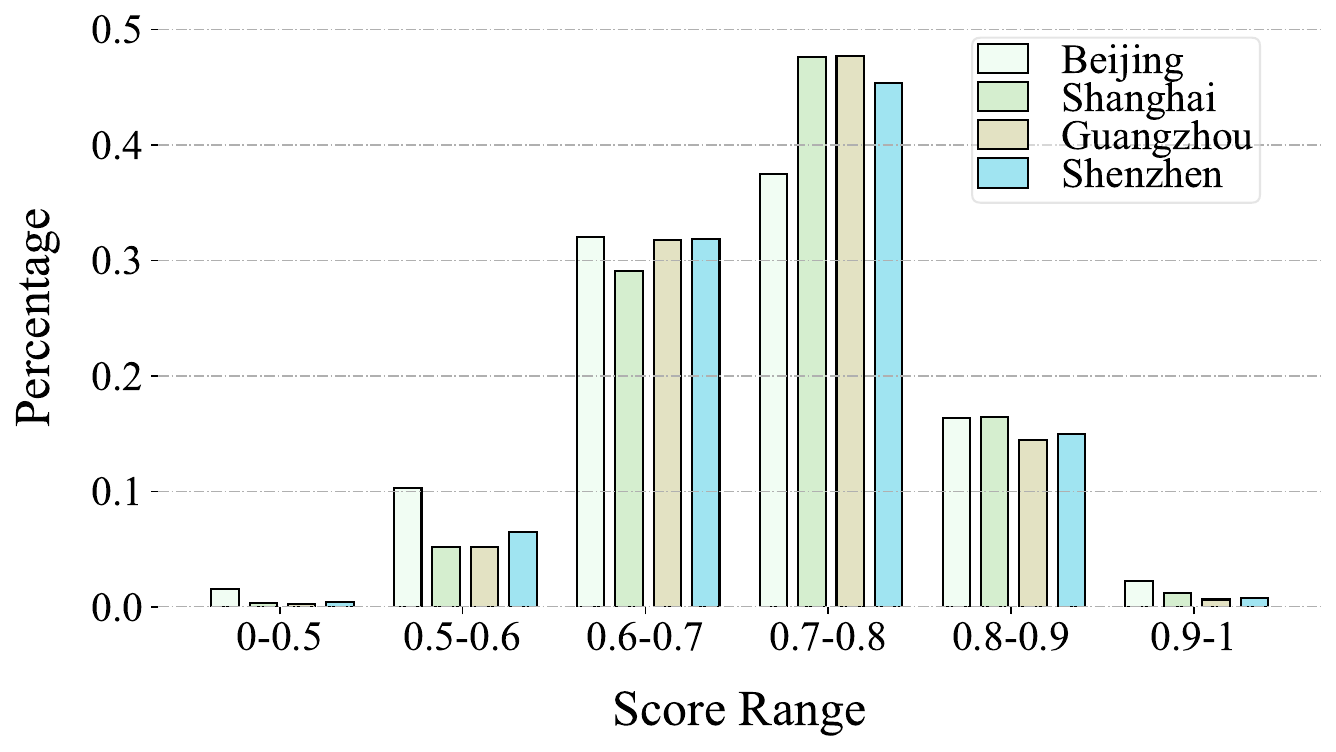}
    \caption{\texttt{PerceptionScore} distribution on $\mathtt{\dataset}$ dataset for 4 representative cities.}
    \label{fig:perceptionscore}
    \vspace{-2em}
\end{figure}

In our paper, we select the state-of-the-art LMM ShareGPT4V, which demonstrates performance comparable to GPT4V in image captioning. 
Here, we compute the distribution of \texttt{PerceptionScore} for descriptions generated by ShareGPT4V in Figure \ref{fig:perceptionscore}. 
Moreover, in our proposed dataset, we filter out the texts with \texttt{PerceptionScore} lower than 0.6. To validate the effectiveness
of this Text Calibration process, we report the performance of using raw generated
descriptions (i.e., UrbanVLP w/o Text Calibration) for comparison.


Table \ref{tab:ablationresults_beijing} demonstrates that UrbanVLP consistently surpasses this variant across various indicators. This outcome confirms the effectiveness of our Text Calibration method,
suggesting that aligning more relevant and noise-free textual information with image features enhances coherence and meaning in visual representation.

\begin{table}[h!]
\centering
\caption{UrbanVLP performance with and without Text Calibration. TC stands for Text Calibration.}
\setlength{\tabcolsep}{0.8mm}
\scalebox{0.7}{
\begin{tabular}{l|ccc|ccc|ccc}
\hline
\multirow{2}{*}{\textbf{Model}}
& \multicolumn{3}{c|}{\textbf{Carbon}} & \multicolumn{3}{c|}{\textbf{Population}} & \multicolumn{3}{c}{\textbf{GDP}} \\

& $\text{R}^2$ & $\text{RMSE}$ & $\text{MAE}$ & $\text{R}^2$ & $\text{RMSE}$ & $\text{MAE}$ & $\text{R}^2$ & $\text{RMSE}$ & $\text{MAE}$ \\
\hline

UrbanVLP w/o TC & 0.724 & 0.542 & 0.476 & 0.686 & 0.559 & 0.433 & 0.501 & 0.697 & 0.436 \\
UrbanVLP & \textbf{0.769} & \textbf{0.477} & \textbf{0.369} & \textbf{0.714} & \textbf{0.523} & \textbf{0.411} & \textbf{0.537} & \textbf{0.684} & \textbf{0.416} \\
\hline
\end{tabular}
}
\label{tab:ablationresults_beijing}
\vspace{-1em}
\end{table}

%% file: appendix/practicality.tex
\section{K ~~~Practicality}
\label{appendix:rq4}

\begin{figure*}[t!]
    \centering
    \includegraphics[width=0.95\linewidth]{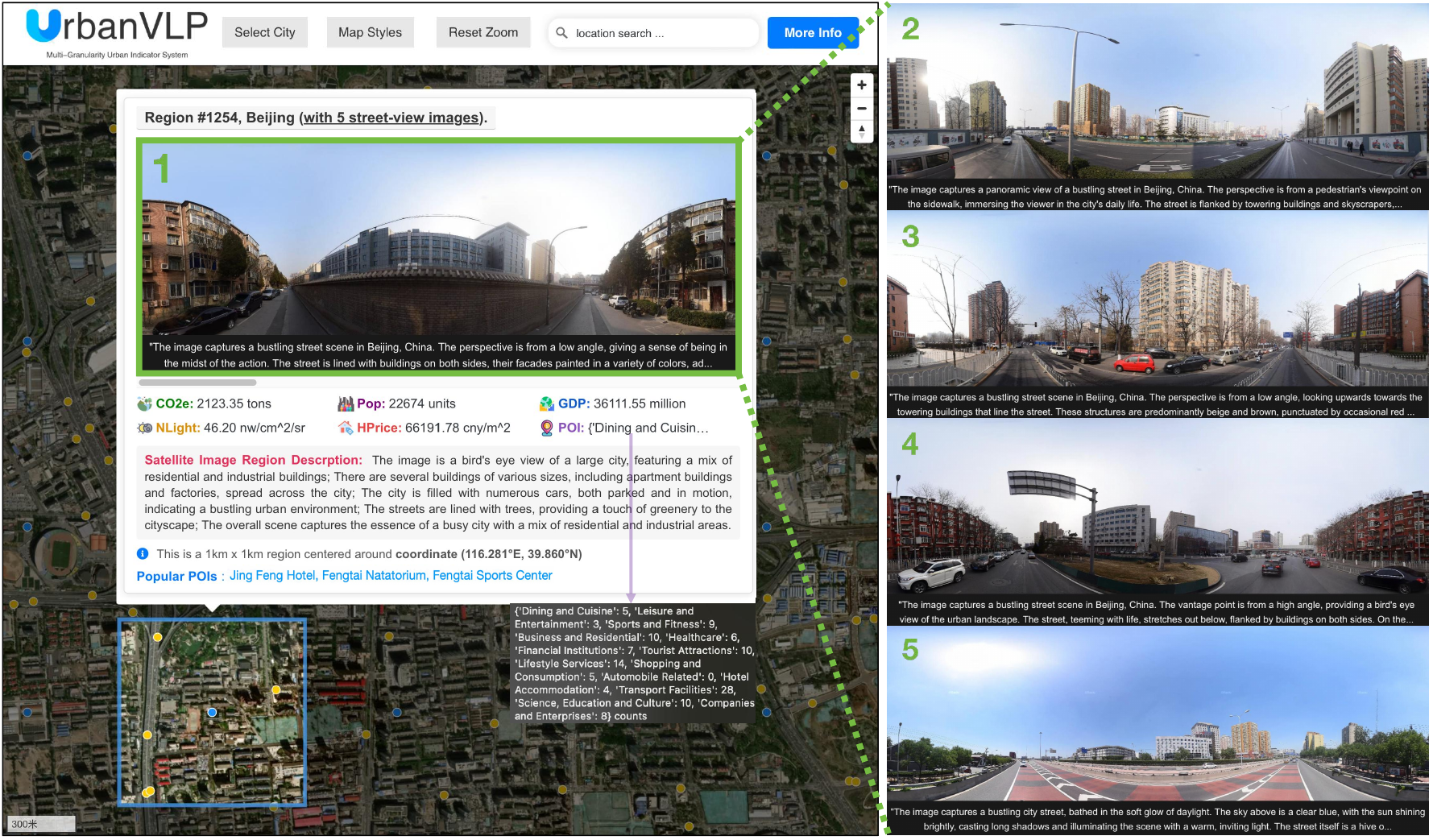} 
    \caption{\model system.}
    \label{fig:system}
\end{figure*}

To demonstrate the practical applications of our model in real-life scenarios, we have developed a system bearing the same name \model.
The \model system is a novel web-based system developed to integrate LMMs with a multi-scale urban socioeconomic indicator framework using the Mapbox platform \cite{mapbox}.  Figure \ref{fig:system} showcases the user-friendly interface of the map, highlighting functionalities that facilitate zooming, searching for locations, and navigating through various sectors.

Each urban region, captured by a satellite image and covering a 1km $\times$ 1km area, is marked by a blue dot. Yellow points surround the blue dot, indicating the locations of street-view images accessible within that region. Engaging with these points allows users to uncover a suite of urban metrics, such as carbon emissions, population, GDP, night light, house price, and POI. The system's visual components are augmented by a captioning feature that provides clear and informative textual descriptions, offering insights into the spatial characteristics of the designated area.

Furthermore, a street-view panel allows users to peruse individual street-view images alongside pertinent captions. The system also facilitates the discovery of popular POIs within any given area, enriching the understanding of its functional characteristics. Ultimately, \model is designed to distill complex urban data into a visual and user-friendly format, thereby offering a holistic and detailed view of urban landscapes and their key indicators.
We have also prepared a demonstration video in the supplementary material to showcase the utility of our system.

%% file: appendix/limitation_and_social_impact.tex
\section{L ~~~Limitations and Social Impact}
Due to the easier accessibility of the data, our work primarily focuses on Chinese first-tier cities, which introduces a certain bias and can be regarded as a limitation of our study. 
Moreover, the model's performance is significantly influenced by image resolution. Higher quality image data is required to achieve optimal performance.
Additionally, the inability to automatically correct low-quality text highlights an area for future improvement.

Our work can be deployed in practical applications, as demonstrated by the example provided in the paper, thereby aiding in urban planning and decision-making processes. However, it is crucial to note that satellite and street-view images might inadvertently contain private information. 
The street-view images we collected have been blurred to obscure relevant information, such as license plates, thereby significantly reducing associated risks. Nevertheless, we strongly advise users to perform a comprehensive analysis of the data prior to deploying our model in practical contexts.